\documentclass[review]{elsarticle}

\usepackage{lineno,hyperref}
\usepackage{booktabs}
\usepackage{amsmath}
\usepackage{amsfonts}
\usepackage{mathrsfs}
\usepackage{dsfont}
\usepackage{amsthm}

\usepackage{setspace}
\usepackage{float}
\modulolinenumbers[5]
\usepackage{color}
\usepackage[numbers]{natbib}

\usepackage[export]{adjustbox}
\usepackage{xcolor}
\usepackage{tikz}
\usepackage{graphicx}
\usepackage{multirow}
\usepackage{bbm} 
\usepackage{enumitem}

\usepackage{subcaption}
\usepackage{caption}
\usepackage{algorithm}
\usepackage{algorithmicx}
\usepackage{algpseudocode}
\usepackage{bm}
\usepackage{bbm}
\usepackage{indentfirst}
\usepackage{url}

\usepackage{hyperref}
\usepackage{cleveref}

\journal{Journal of \LaTeX\ Templates}

\begin{document}

\begin{frontmatter}

\title{Diverse Teacher-Students for Deep Safe Semi-Supervised Learning under Class Mismatch}





\author[mymainaddress]{Qikai Wang\corref{cor1}}
\author[mymainaddress]{Rundong He\corref{cor1}}
\author[mymainaddress]{Yongshun Gong\corref{mycorrespondingauthor}}
\author[mymainaddress1]{Chunxiao Ren}

\author[mymainaddress]{Haoliang Sun}
\author[mymainaddress3]{Xiaoshui Huang}
\author[mymainaddress]{Yilong Yin\corref{mycorrespondingauthor}}

\cortext[cor1]{These authors contributed equally to this work}
\cortext[mycorrespondingauthor]{Corresponding authors}

\address[mymainaddress]{School of Software, Shandong University}
\address[mymainaddress1]{Shandong International Talent Exchange Service Center}

\address[mymainaddress3]{Shanghai Artificial Intelligence Laboratory}

\begin{abstract}
Semi-supervised learning can significantly boost model performance by leveraging unlabeled data, particularly when labeled data is scarce. However, real-world unlabeled data often contain unseen-class samples, which can hinder the classification of seen classes. To address this issue, mainstream safe SSL methods suggest detecting and discarding unseen-class samples from unlabeled data. Nevertheless, these methods typically employ a single-model strategy to simultaneously tackle both the classification of seen classes and the detection of unseen classes. Our research indicates that such an approach may lead to conflicts during training, resulting in suboptimal model optimization. Inspired by this, we introduce a novel framework named Diverse Teacher-Students (\textbf{DTS}), which uniquely utilizes dual teacher-student models to individually and effectively handle these two tasks. DTS employs a novel uncertainty score to softly separate unseen-class and seen-class data from the unlabeled set, and intelligently creates an additional ($K$+1)-th class supervisory signal for training. By training both teacher-student models with all unlabeled samples, DTS can enhance the classification of seen classes while simultaneously improving the detection of unseen classes. Comprehensive experiments demonstrate that DTS surpasses baseline methods across a variety of datasets and configurations. Our code and models can be publicly accessible on the link \href{https://github.com/Zhanlo/DTS}{\small\url{https://github.com/Zhanlo/DTS}}.
\end{abstract}


\end{frontmatter}


\section{Introduction}\label{sec:introduction}
In the realm of deep learning, the existence of high-quality, large-scale labeled datasets has significantly accelerated the evolution and refinement of algorithms \cite{DBLP:conf/cvpr/DengDSLL009, DBLP:conf/cvpr/HeCXLDG22, DBLP:conf/cvpr/He0WXG20, DBLP:conf/nips/KrizhevskySH12, DBLP:conf/eccv/LinMBHPRDZ14, DBLP:conf/iclr/RizveDRS21}. However, this has also led to a heavy reliance on such datasets. Objectively, the procurement of these extensive, high-quality labeled datasets often comes at a substantial cost, to some extent, which hinders the rapid advancement of algorithmic development. Amidst this backdrop, by propagating the semantic information from labeled data into vast amounts of unlabeled data \cite{DBLP:conf/aaai/Cascante-Bonilla21, CHEN2024108367, DBLP:conf/cvpr/IscenTAC19, MAWULI2023119235, DBLP:conf/nips/OliverORCG18, DBLP:conf/nips/RenYS20, DBLP:conf/eccv/WangLY22, DBLP:conf/cvpr/XieLHL20}, semi-supervised learning allows for effective model training, thereby mitigating the algorithm's heavy dependency on massive labeled datasets.

\begin{figure}[t]
\centering
\includegraphics[width=\columnwidth]{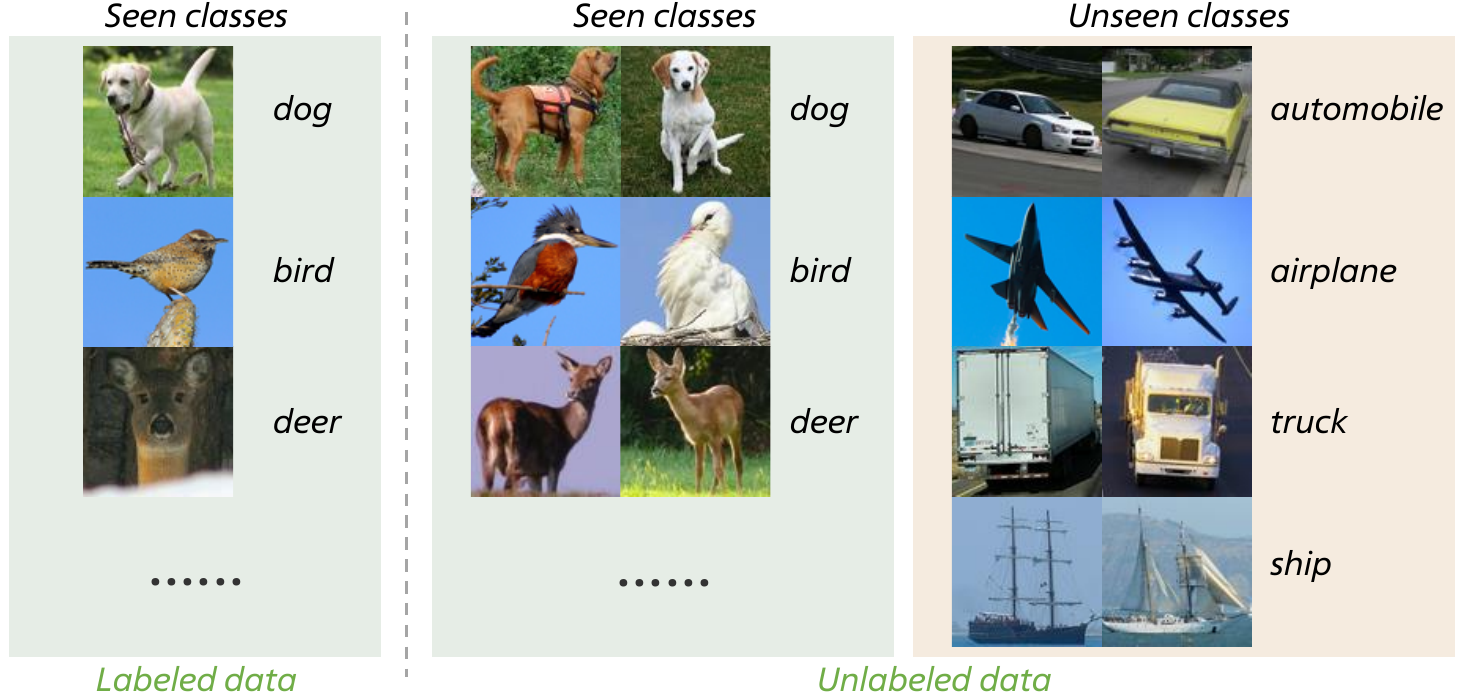}
\caption{Illustration of the safe semi-supervised learning setting with class distribution mismatch. Take the categories of CIFAR-10 dataset as an example, in which the animal categories are seen classes and the other categories are unseen classes.}
\label{fig of setting}
\end{figure}

Nevertheless, most existing semi-supervised learning approaches rest on a fundamental assumption that labeled and unlabeled datasets share the same semantic space. However, this assumption is often unrealistic, as effectively screening and filtering diverse sources of unlabeled data under budget limitations and resource constraints is a formidable task. Further complicating this scenario is the presence of unseen-class samples within the training data, which can significantly disrupt the learning process, potentially leading to unstable outcomes or severe performance degradation \cite{DBLP:conf/aaai/ChenZLG20, DBLP:conf/icml/GuoZJLZ20, DBLP:conf/nips/OliverORCG18}. Several similar definitions have emerged to describe this scenario, including safe SSL \cite{DBLP:conf/icml/GuoZJLZ20}, open-set SSL \cite{DBLP:conf/iccv/00100S023, LIU2024103814, openmatch, DBLP:conf/eccv/YuIIA20}, and the challenge of managing UnLabeled data from Unseen Classes in Semi-Supervised Learning (ULUC-SSL) \cite{DBLP:journals/tkde/HeHLY24}. In this paper, we prefer to refer to it as the safe classification problem of safe SSL \cite{DBLP:conf/icml/GuoZJLZ20}, to emphasize that our core goal is to ensure that the model's performance is not compromised, or even degraded, by the presence of unseen-class samples during the training process.

\begin{figure}[t]
\centering
\includegraphics[width=\columnwidth]{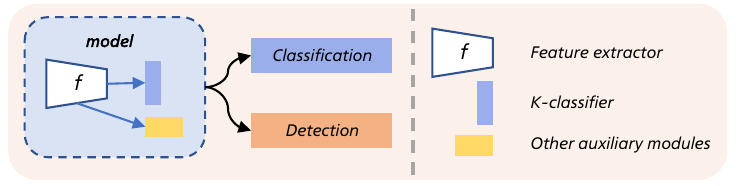}
\caption{A common safe semi-supervised learning framework. Such frameworks generally train a singular model to undertake dual tasks: classification of seen-classes and detection of unseen-classes.}
\label{fig:common_model_framework}
\end{figure}

More precisely, as shown in Figure \ref{fig of setting}, safe SSL is concerned about the semi-supervised learning scenarios with class distribution mismatch, where the unlabeled data includes samples from both seen and  unseen classes \cite{DBLP:conf/cvpr/HeHLY22, DBLP:journals/tkde/HeHLY24}. Specifically speaking, seen classes refer to categories that are known and labeled during training, whereas unseen classes denote those not labeled or explicitly identified during this phase. A key factor in this context is the \textbf{class mismatch ratio}, which measures the proportion of seen to unseen classes within the unlabeled set. This ratio significantly influences the complexity and challenge of the SSL task. A high class mismatch ratio suggests a predominant presence of unseen classes relative to seen classes in the unlabeled data, complicating the learning process. It requires methods to identify and manage these unseen classes during training, even though they are not the target of the final classification tasks.

Although this challenge is prevalent in various real-world applications, it remains relatively underexplored in current research. Many real-world challenges present cases where unlabeled data includes instances from unseen-classes. For instance, in the field of digital medical imaging, continuous advancements often lead to the discovery of new disease types or novel radiological presentations of existing diseases. Consequently, novel disease types might infiltrate the unlabeled medical imaging data, which were not present in the original training sets. These unseen-class instances can significantly compromise the accuracy and reliability of diagnostic models. If the models fail to recognize these new examples of unseen classes and appropriately flag them, the resultant errors may endanger patient lives, as dramatically illustrated by the emergence of COVID-19 \cite{DBLP:journals/tmi/HanWHLCZWZ20}.

Faced with the potential disturbances that samples from unseen classes might introduce during model training, previous strategies, such as UASD \cite{DBLP:conf/aaai/ChenZLG20}, has adopted a straightforward approach: detect and remove these unseen-class samples, focusing only on the reliable seen-class samples within the unlabeled data for training. However, as shown in Figure \ref{fig:common_model_framework}, although these methods may contain other auxiliary modules to help with training, most of them only resort to a singular model to accomplish two distinct tasks: classification of seen-classes and detection of unseen-classes. Upon a deeper analysis, we identify three subtle yet significant issues inherent to this approach:

First, mainstream safe SSL methods adopt the optimization strategy of training a single model, typically using the classifier for seen classes to detect unseen classes \cite{DBLP:conf/cvpr/HeHLY22, DBLP:journals/tkde/HeHLY24, DBLP:conf/nips/SohnBCZZRCKL20}. However, during the utilization of unlabeled data, unseen-class samples may be mislabeled as seen ones and vice versa, causing confusion in distinguishing between the two. Moreover, this strategy prioritizes seen-class accuracy, overlooking the importance of detecting unseen classes. Over iterations, while seen-class classification may improve, the model's ability to identify unseen classes weakens. This leads to an accumulated bias toward uncertainty and suboptimal use of unlabeled data, which ultimately hampers the overall classification performance. 

Second, current methods, when filtering seen and unseen-class samples from unlabeled data, frequently depend on multiple hyperparameters. These hyperparameters usually require prior knowledge and necessitate distinct adjustments across varying tasks and datasets, complicating their practical application. 

Third, unseen-class samples in unlabeled data are not fully utilized. Many methods adopt the strategy of detecting and removing these samples, which does not fully harness their potential, as they are not treated as a distinct class and incorporated into supervised training.

Given the above challenges, we propose the Diverse Teacher-Students (\textbf{DTS}) framework. DTS employs two independent teacher-student models: one concentrating on seen-class classification and another dedicated to unseen-class detection. This segregated strategy eliminates interference between tasks, facilitating concurrent optimization of both. Moreover, we introduce a soft-weighting module capable of subtly soft-separating unseen-class samples within unlabeled data. Through the soft-weighting module, our approach does not necessitate the acquisition of extra prior knowledge specific to each dataset. This simplifies the application process and enhances the feasibility of our method across diverse datasets. Ultimately, DTS treats the identified unseen-class data as the ($K$+1)-th class and trains using a $K+1$ classifier, maximizing the potential of unseen-class data while preventing disruption to the original $K$ classes classification.

Our contributions can be summarized as follows:
\begin{itemize}[noitemsep,topsep=0pt,leftmargin=*]
    \item We identify that mainstream deep safe SSL methods have a notable limitation: they use a single model for both seen-class classification and unseen-class detection, hindering optimal performance.
    
    \item We introduce an innovative  framework called \textbf{DTS}. Unlike mainstream methods, DTS employs two distinct models to independently address the two core tasks, circumventing interference between them.
    
    \item We introduce a novel soft-weighting module that utilizes an innovative uncertainty score to softly separate unseen-class samples from unlabeled data, ingeniously harnessing all unlabeled data for training.
    
    \item Extensive experiments demonstrate the superiority of DTS in deep safe SSL tasks, highlighting its dual advantage in enhanced classification performance and stable detection of unseen classes.
\end{itemize}

\section{Related Work}
\subsection{Deep Semi-Supervised Learning}
Semi-supervised learning holds immense potential within the realm of machine learning \cite{DBLP:journals/ml/EngelenH20, DBLP:journals/tnn/ChapelleSZ09, DBLP:conf/nips/WangCFSTHWY0GQW22, DBLP:conf/iccv/YangZQQSZ23, DBLP:journals/corr/abs-2103-00550, DBLP:conf/cvpr/ZhengYHWQX22}. Compared to supervised learning, it escapes the shackles of being entirely reliant on labeled data, shifting its focus towards the rich sample characteristics and structural information concealed within unlabeled data. In contrast with unsupervised learning, it successfully propagates the semantic information from labeled data to unlabeled counterparts, thereby streamlining its application in downstream tasks.

Self-training and consistency regularization stand out as cornerstone techniques of contemporary deep semi-supervised learning \cite{DBLP:conf/iclr/LaineA17, DBLP:conf/iccv/0001XH21, DBLP:conf/iccv/00100S023, DBLP:conf/nips/SajjadiJT16, DBLP:conf/nips/TarvainenV17, DBLP:conf/cvpr/WangQLSZC23}. In particular, the pseudo-labeling technique within self-training plays a pivotal role. It iteratively enlarges the ``labeled data" by speculatively assigning pseudo-labels to unlabeled samples, enabling unlabeled data to generate supervisory signals for model training \cite{lee2013pseudo, DBLP:conf/iclr/WangJWWM23}. This strategy has become almost inescapable in current semi-supervised learning paradigms. Consistency regularization, on the other hand, operates under the presumption that training samples share identical labels with their proximate synthetic counterparts (or similar views). By disseminating the labels from seen-class samples and confident predictions of unseen-class samples, it harnesses unlabeled data even more effectively \cite{DBLP:conf/nips/BerthelotCGPOR19, DBLP:journals/pami/MiyatoMKI19, openmatch, DBLP:conf/nips/SohnBCZZRCKL20}. For example, FixMatch \cite{DBLP:conf/nips/SohnBCZZRCKL20} is an efficient approach that capitalizes on both techniques. It refines consistency regularization through the distinct deployment of weak and strong augmentations. And there are many other works that have made important technical contributions to the research of SSL. ReMixMatch \cite{DBLP:conf/iclr/BerthelotCCKSZR20} introduces distribution alignment and augmentation anchoring. FlexMatch \cite{zhang2021flexmatch} and FreeMatch \cite{wang2022freematch} propose to introduce currency pseudo labeling and self-adaptive threshold to deal with the learning difficulty of different classes.

Despite the promising strides in SSL, there are inherent limitations that need to be addressed. The most salient among these is the prevalent assumption that labeled and unlabeled data samples share identical category spaces. When this assumption is breached, there is a tangible dip in the performance of seen-class classification.

\subsection{Safe Semi-Supervised Learning}
To address the issues mentioned above, mainstream safe SSL strategies tend to detect and exclude unseen-class samples to avoid misunderstandings in the training process. For example, UASD \cite{DBLP:conf/aaai/ChenZLG20} relies on the confidence threshold of model predictions to identify outliers. In contrast, Safe-Student \cite{DBLP:conf/cvpr/HeHLY22} utilizes the logit output of the model to define an energy difference score, thereby surpassing the straightforward confidence-based method and providing a more precise distinction between seen and unseen-class samples. In addition, there are some other excellent methods. MTCF \cite{DBLP:conf/eccv/YuIIA20} alternates the update of network parameters and the OOD score using a joint optimization framework. TOOR \cite{DBLP:journals/tmm/Huang0023} considers OOD data closely related to ID and labeled data as recyclable, enriching information for secure SSL using both ID and "recyclable" OOD data. Building upon Fixmatch, OpenMatch \cite{openmatch} innovatively incorporates one-vs-all (OVA) classifiers to detect unseen-class samples, achieving promising results. IOMatch \cite{DBLP:conf/iccv/00100S023}, on the other hand, accomplishes open-set classification better by leveraging K-classifier, ($K$+1)-classifier, and one-vs-all (OVA) classifiers to produce unified Open-Set targets.

Most of existing methods, however, often overly focus on optimizing the performance of seen-class classification, neglecting the detection capability for unseen-classes during the self-training phase. Inevitably, some unseen-class samples get misclassified as seen ones and vice versa. This misclassification leads to a bias of uncertainty, potentially causing the model to increasingly confuse between seen and unseen classes, as we discussed in the Introduction. To address this vital problem, we advocate for the decoupling of the two primary tasks: seen-class classification and unseen-class detection. By optimizing them concurrently, we aim to prevent the significant accumulation of uncertainty bias and ensure maximal utilization of unlabeled data.

\section{Methodology}
\subsection{Learning Set-Up}

\begin{figure*}[t]
\centering
    {\includegraphics[width=\textwidth]{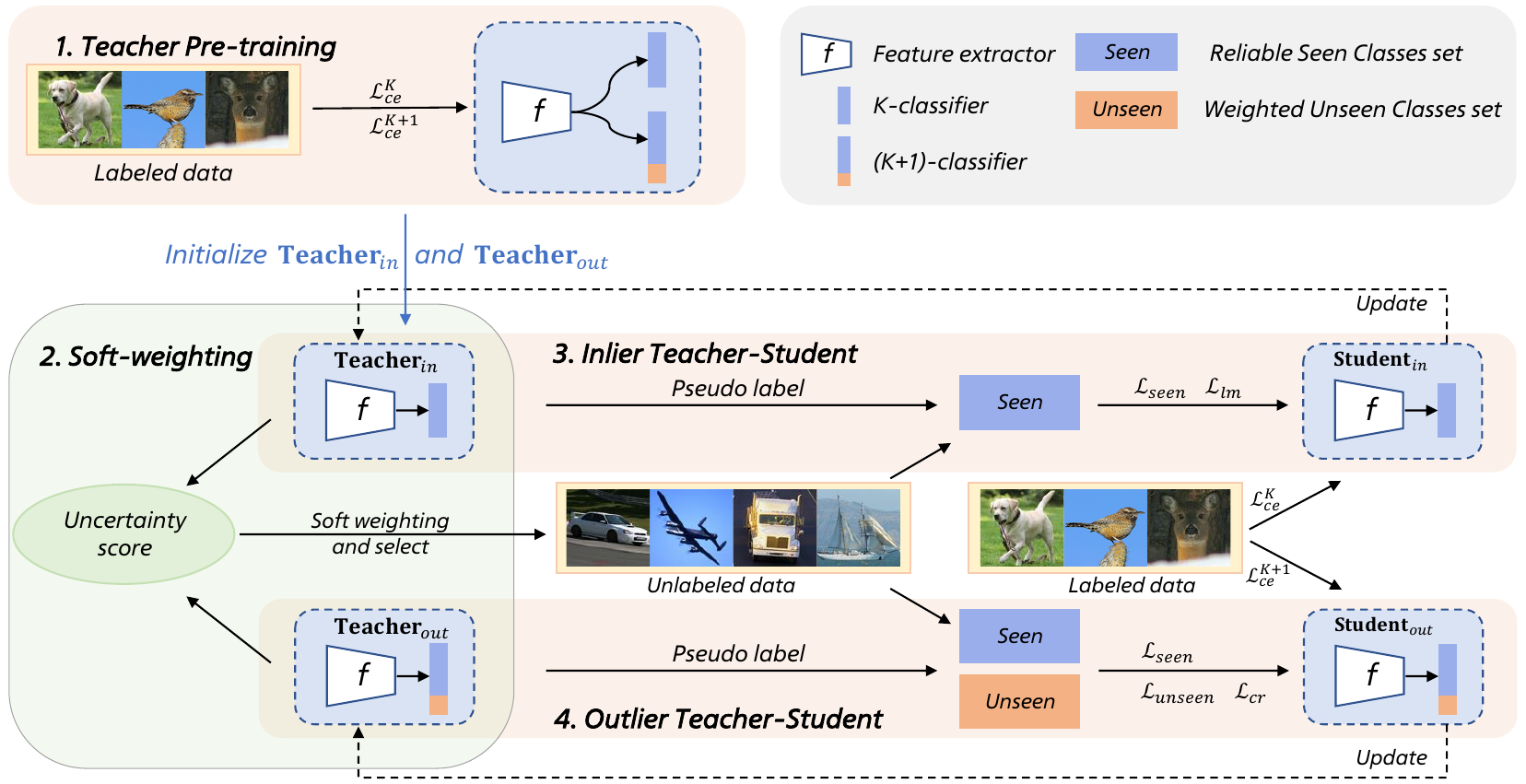}}
\caption{Overview of DTS. DTS offers a robust approach to deep, safe semi-supervised learning for datasets with unseen classes. It integrates four pivotal modules, starting with a pre-trained teacher model to establish the Inlier Teacher-Student (ITS) and Outlier Teacher-Student (OTS) frameworks. Each iteration advances with the student model refining the teacher model. ITS and OTS then engage a soft-weighting mechanism to capitalize on unlabeled data, achieving a harmonized optimization of the training objectives.}
\label{fig of Overview}
\end{figure*}

Let $L=\{(\boldsymbol {x}_1,y_1),(\boldsymbol {x}_2,y_2),...,(\boldsymbol {x}_m,y_m)\}$ represent the labeled dataset. Here, m signifies the quantity of labeled samples. For any $i \in [1,2,\cdots,m]$, $\boldsymbol {x}_i \in \mathbb R^d$ is the representation of a labeled sample where d is the number of input dimension , and $y_i\in\{1,2,...,K\}$ designates its corresponding label. On the other hand, $U=\{\boldsymbol {u}_1,\boldsymbol {u}_2,...,\boldsymbol {u}_n\}$ represents the unlabeled dataset. In this set, n specifies the number of unlabeled samples, and each $\boldsymbol{u}_i$ denotes an unlabeled sample, with the condition that m$\ll$n. Notably, the class of $\boldsymbol{u}_i$ might not align with any of the predefined $K$ seen classes, we consider these samples of unseen classes as a single class: the (\textit{K}+1)-th class. Additionally, during the model training process, we employ data augmentation strategies to produce perturbed views of the samples. Specifically, $\text{Aug}_{\text{strong}}(\cdot)$ and $\text{Aug}_{\text{weak}}(\cdot)$ represent strong and weak data augmentation processes, respectively.

The framework of DTS, as depicted in Figure \ref{fig of Overview}, comprises four modules: (1) The Teacher Pre-training Module, which trains a single teacher model equipped with two classifiers using all labeled data, setting the stage for the subsequent teacher-student frameworks. (2) The Soft Weighting Module, using two teacher models, introduces a novel uncertainty score that assesses the uncertainty of each unlabeled sample, thereby realizing the soft separation of seen-class samples and unseen-class samples. (3) The Inlier Teacher-Student (ITS) Module features a specialized teacher-student framework for classifying seen classes, utilizing the feature extractor and the $K$-classifier from the pre-trained teacher model for its initial setup. (4) In a similar vein, the Outlier Teacher-Student (OTS) Module, dedicated to detecting unseen classes, also bases its initial configuration on the feature extractor and the (\textit{K}+1)-classifier of the pre-trained teacher model.

\subsection{Teacher Pre-Training}
We initiate our process by pre-training the teacher model using all available labeled data. Given a labeled batch, we denote it as $\mathcal{X} = \{ (\boldsymbol{x}_i, y_i) \mid i = 1, \ldots, B \}$, where $B$ is the batch size.

As illustrated in Figure \ref{fig of Overview}, during the pre-training phase, the feature extractor $f(\cdot)$ transforms these samples into a $D$-dimensional feature space, expressed as:$\quad h_i=f(\text{Aug}_{\text{strong}}(x_i))\in\mathbb{R}^D$. Following this, two distinct classifiers $\Phi$ and $\psi$ map these feature vectors into the predicted $K$-dimensional and ($K$+1)-dimensional probability distribution, respectively. The relationship can be mathematically captured by:
\begin{equation}
\begin{aligned}
&\boldsymbol p_i={\Phi}(h_i)\in\mathbb{R}^K, \\
&\boldsymbol p_i'=\psi(h_i)\in\mathbb{R}^{K+1}.
\end{aligned}
\tag{1}
\end{equation}

In the final step, although the labeled set does not contain samples of the ($K$+1)-th class, we utilize the cross-entropy losses of the two classifiers to optimize the teacher model and complete the pre-training initialization. This process is formalized as follows:
\begin{equation}
\begin{aligned}
&\mathcal{L}_{ce}^{K}(\mathcal{X})=\frac{1}{B}\sum_{i=1}^{B}\text{H}(y_{i},\boldsymbol{p}_{i}), \\
&\mathcal{L}_{ce}^{K+1}(\mathcal{X})=\frac{1}{B}\sum_{i=1}^{B}\text{H}(y_{i},\boldsymbol{p}_{i}').
\end{aligned}
\tag{2}
\end{equation}

We use the sum of two losses to optimize the teacher model: $\mathcal L_{pretrain}=\mathcal{L}_{ce}^{K}+\mathcal{L}_{ce}^{K+1}$. In Algorithm \ref{alg:algorithm1}, we present the details of the pre-training process for the teacher model and how ITS and OTS are initialized post-pre-training.

\begin{algorithm*}[t]
\footnotesize
\caption{Pre-training of the \textbf{Teacher} model}
\label{alg:algorithm1}
\textbf{Setup}:

\quad Pre-train epochs = 1000 (for the CIFAR-10 dataset). $\text{Aug}_{\text{weak}}(\cdot)$ and $\text{Aug}_{\text{strong}}(\cdot)$: Weak and strong augmentation. $f(\cdot)$: Feature extractor. $\Phi$: The K-classifier. $\psi$: The ($K$+1)-classifier.

\quad \textbf{Teacher}: The pre-trained teacher model. 

\quad $\textbf{Teacher}_{in}$: The teacher model for seen-class classification. 

\quad $\textbf{Student}_{in}$: The student model for seen-class classification. 

\quad $\textbf{Teacher}_{out}$: The teacher model for unseen-class detection. 

\quad $\textbf{Student}_{out}$: The student model for unseen-class detection. 

\textbf{Input}: 

\quad Labeled batch: $\mathcal{X} = \{ (\boldsymbol{x}_i, y_i) \mid i \in \{1, \ldots, B\} \}$.

\begin{algorithmic}[1] 
\State $h_i = f(\text{Aug}_{\text{strong}}(\boldsymbol{x}_i))$
\Comment{\textcolor{gray}{\textit{Obtain the features of the labeled samples.}}}
\State $\boldsymbol{p}_i = \Phi(h_i)$, $\boldsymbol{p}_i^{\prime} = \psi(h_i)$
\Comment{\textcolor{gray}{\textit{Get $K$-classes and $K$+1 classes predictions.}}}
\State $\mathcal{L}_{ce}^{K}(\mathcal{X}) = \frac{1}{B} \sum_{i=1}^{B} \text{H}(y_i, \boldsymbol{p}_i)$
\Comment{\textcolor{gray}{\textit{Calculate the $K$-classification loss.}}}
\State $\mathcal{L}_{ce}^{K+1}(\mathcal{X}) = \frac{1}{B} \sum_{i=1}^{B} \text{H}(y_i, \boldsymbol{p}_i^\prime)$
\Comment{\textcolor{gray}{\textit{Calculate the ($K$+1)-classification loss.}}}

\end{algorithmic}
\textbf{Output}: The overall loss $\mathcal{L}_{pretrain} = \mathcal{L}_{ce}^{K} + \mathcal{L}_{ce}^{K+1}$ to update the \textbf{Teacher} model parameters.

\textbf{Model initialization}: After the pre-training process, we use the \textbf{Teacher} model to initialize these models:

\quad $\textbf{Teacher}_{in} = \textbf{Student}_{in} : f(\cdot) + \Phi$

\quad $\textbf{Teacher}_{out} = \textbf{Student}_{out} : f(\cdot) + \psi$

\end{algorithm*}

\subsection{Soft-Weighting}
\label{soft-weighting}
Current Deep Safe SSL methods often set confidence-based thresholding on the model's predictions to process unlabeled data \cite{DBLP:conf/aaai/ChenZLG20, DBLP:conf/nips/SohnBCZZRCKL20}. They discard uncertain samples and allocate pseudo-labels to those with high confidence to differentiate between seen and unseen-class samples. The limitation of such strategies is that they do not make full use of the potential within the unlabeled data. Moreover, solely relying on $K$-classifier to detect unseen samples is suboptimal.

Leveraging two teacher models as a foundation, we propose an innovative uncertainty score. We employ both $K$-classifier and ($K$+1)-classifier to effectively identify unseen-class samples within the unlabeled data. Specifically, for an unlabeled sample $\boldsymbol u_{i} \in U$, its uncertainty score is given by:
\begin{equation}
\begin{aligned}
&\boldsymbol{p}_i^{ITS}={\Phi}(f(\text{Aug}_{\text{weak}}(\boldsymbol{u}_i))) \in\mathbb{R}^K,
\\
&\boldsymbol{p}_i^{OTS}={\psi}(f(\text{Aug}_{\text{weak}}(\boldsymbol{u}_i))) \in\mathbb{R}^{K+1},
\\
&s(\boldsymbol{u}_i) = \gamma \cdot \left(1 - \max(\boldsymbol{p}_i^{ITS})\right) + (1-\gamma) \cdot \boldsymbol{p}_i^{OTS}[-1],
\end{aligned}
\tag{3}
\end{equation}
where $\gamma \in [0,1]$ is a balanced weight. $\boldsymbol{p}_i^{ITS} \in \mathbb{R}^K$ and $\boldsymbol{p}_i^{OTS} \in \mathbb{R}^{K+1}$ represent the predicted $K$-dimensional and ($K$+1)-dimensional probability distribution outputted by the $\textbf{Teacher}_{in}$ and $\textbf{Teacher}_{out}$ models, respectively. Considering each unlabeled sample as a potential member of unseen classes, we construct a softly weighted set, $U_{unseen}$, of these samples, with their uncertainty scores serving as weights:
\begin{equation}
\begin{aligned}
U_{unseen} = \{ (\boldsymbol{u}_i, w_i) | i \in\{1,\cdots,n\}, \boldsymbol{u}_i \in U \text{ and } w_i = s(\boldsymbol{u}_i) \}.
\end{aligned}
\tag{4}
\end{equation}

Within the set $U_{unseen}$, we classify each sample as belonging to the ($K$+1)-th class.  This approach allows for a soft separation of unseen-class samples from the unlabeled data. In this manner, we can achieve effective detection of unseen classes, thereby enabling more efficient utilization of the unlabeled set and enhancing model performance. Moreover, this is achieved without the need for any prior knowledge specific to the dataset or hyperparameters, making the task simpler, more feasible, and convenient to apply.

\subsection{Inlier Teacher-Student}
As previously discussed, the Inlier Teacher-Student module is specifically designed to handle the classification of seen classes. To begin with, similar to the pre-training of the teacher model, we employ labeled data to compute the cross-entropy loss $\mathcal L_{ce}^{K} (\mathcal X)$ for training the $\textbf{Student}_{in}$. 

For learning from the seen-class samples in the unlabeled data, we rely on the traditional confidence-based thresholding and the uncertainty score, which are simple but effective enough. Specifically, for a given unlabeled batch $\mathcal{U} = \{ (\boldsymbol{u}_i) : i \in (1, \ldots, \mu B) \}$, where $\mu$ is a hyperparameter that defines the relative sizes of $\mathcal{X}$ and $\mathcal{U}$, let $\boldsymbol{p}_i^{ITS}=\Phi(f(\text{Aug}_{\text{weak}}(\boldsymbol{u}_i))) \in \mathbb{R}^K$ represents the predicted probability distribution in the $K$-dimensional space for $\boldsymbol u_i$ as output by the $\textbf{Teacher}_{in}$ model:
if $\max(\boldsymbol{p}_i^{ITS}) > threshold$ and $\max(\boldsymbol{p}_i^{ITS}) > s(\boldsymbol{u}_i)$, then $\boldsymbol u_i$ is considered as a trustworthy seen-class sample. The class $\tilde{y}_i^u$, corresponding to $\operatorname{argmax}(\boldsymbol{p}_i^{ITS})$, is adopted as the sample's pseudo-label to guide the training of the $\textbf{Student}_{in}$ model. This process is formulated as:
\begin{equation}
\begin{aligned}
&\boldsymbol{p}_i={\Phi}(f(\text{Aug}_{\text{strong}}(\boldsymbol{u}_i))) \in\mathbb{R}^K,
\\
&\mathcal{F}(\boldsymbol{u}_i)=\mathbbm{1}(\max(\boldsymbol{p}_i^{ITS}) > \tau) \cdot \mathbbm{1} (\max(\boldsymbol{p}_i^{ITS}) > s(\boldsymbol{u}_i)),
\\
&\mathcal{L}_{seen}(\mathcal U) = \frac1{\mu B} \sum_{i=1}^{\mu B} \text{H}(\tilde{y}_i^u, \boldsymbol{p}_i) \cdot \mathcal{F}(\boldsymbol{u}_i),
\end{aligned}
\tag{5}
\end{equation}
where $\boldsymbol{p}_i \in \mathbb{R}^K$ denote the probability distribution vector in the $K$-dimensional space for $\boldsymbol u_i$ as output by the $\textbf{Student}_{in}$ model, and $\mathcal{F}(\cdot)$ is the filtering function, in which $\mathbbm{1}(\cdot)$ symbolizes the indicator function, $\tau$ is the designated threshold. 

Additionally, to further enhance the classification performance for seen classes, we introduce a Kullback–Leibler (KL) divergence-based loss that allows for robust guidance from $\textbf{Teacher}_{in}$ to $\textbf{Student}_{in}$. For those reliable seen-class samples, the strategy, as a technique validated for its effectiveness within the Safe-Student \cite{DBLP:conf/cvpr/HeHLY22}, encourages the student's logit outputs to closely align with those of the teacher. We refer to this targeted loss as the Logit Matching Loss, defined by the equation:
\begin{equation}
\begin{aligned}
&\boldsymbol p_i=\Phi(f(\text{Aug}_{\text{strong}}(\boldsymbol{u}_i))) \in\mathbb{R}^K,
\\
&\boldsymbol p_i^{ITS}=\Phi(f(\text{Aug}_{\text{weak}}(\boldsymbol{u}_i))) \in\mathbb{R}^K,
\\
&\mathcal{L}_{lm}(\mathcal{U}) = \frac{1}{\mu B} \sum_{i=1}^{\mu B} \mathbb{K}\mathbb{L}(\boldsymbol p_i, \boldsymbol p_i^{ITS})\cdot \mathcal{F}(\boldsymbol{u}_i),
\end{aligned}
\tag{6}
\end{equation}
where $\boldsymbol p_i$ indicates the prediction distribution of the $\textbf{Student}_{in}$ model, $\boldsymbol p_i^{ITS}$ represents the prediction distribution of the $\textbf{Teacher}_{in}$ model, and $\mathbb {KL}$ is the Kullback-Leibler divergence measuring the discrepancy between these distributions.


\subsection{Outlier Teacher-Student}
Similar to Inlier Teacher-student, except Outlier Teacher-student also uses $\mathcal{L}_{ce}^{K+1}(\mathcal{X})$, with the help of $\textbf{Teacher}_{out}$, it also uses the  $\mathcal{L}_{seen}(\mathcal{U})$ to optimize itself.

However, the difference is that Outlier Teacher-student focus more on unseen-class detection in unlabeled data. As mentioned in 
Section \ref{soft-weighting}, the uncertainty score can soft separate unseen-class samples from unlabeled data. These soft-weighted samples are used to calculate the cross-entropy loss function of the ($K$+1)-th class to optimize the detection of unseen classes:
\begin{equation}
\label{equ7}
\begin{aligned}
&\boldsymbol p_i'={\psi}(f(\text{Aug}_{\text{strong}}(\boldsymbol{u}_i))) \in\mathbb{R}^{K+1},
\\
&\mathcal{L}_{unseen}(\mathcal{U})=\frac1{\mu B}\sum_{i=1}^{\mu B} \text{H}(y^{K+1},\boldsymbol p_i')\cdot s(\boldsymbol u_i),
\end{aligned}
\tag{7}
\end{equation}
where $\boldsymbol p_i'$ denotes the predicted probability distribution from $\textbf{Student}_{out}$, $y^{K+1}$ represents the one-hot label vector of the ($K$+1)-th class. 
The supervisory signal for the ($K$+1)-th class, generated through Eq. \ref{equ7}, effectively contributes to the training of the ($K$+1)-th class. This, in turn, enhances the effectiveness of the uncertainty score, further optimizing the model.

Furthermore, we leverage the consistency regularization loss to improve the robustness of $\textbf{Student}_{out}$ model. As illustrated in Eq. \ref{equ8}, for the same sample in the unlabeled batch, we expect the probability distributions obtained from weak augmentation views and strong augmentation views are as similar as possible.
\begin{equation}
\label{equ8}
\begin{aligned}
&\boldsymbol p_{wa}=\psi(f(\text{Aug}_{\text{weak}}(\boldsymbol{u}_i))) \in\mathbb{R}^{K+1},
\\
&\boldsymbol p_{sa}=\psi(f(\text{Aug}_{\text{strong}}(\boldsymbol{u}_i))) \in\mathbb{R}^{K+1},
\\
&\mathcal{L}_{cr}(\mathcal{U})=\frac1{\mu B}\sum_{i=1}^{\mu B}\mathbb{KL}(\boldsymbol p_{wa},\boldsymbol p_{sa}).
\end{aligned}
\tag{8}
\end{equation}

\begin{algorithm*}[t]
\footnotesize
\caption{Optimization of DTS in Every Training Iteration}
\label{alg:algorithm2}
\textbf{Setup}:

\quad Iterations = 3, Train epochs = 400. Total epochs = 3 $\times$ 400 = 1200. $\text{Aug}_{\text{weak}}(\cdot)$ and $\text{Aug}_{\text{strong}}(\cdot)$: Weak and strong augmentation. $f(\cdot)$: Feature extractor. $\Phi$: The K-classifier. $\psi$: The ($K$+1)-classifier. $\tau$: Confidence threshold. $\lambda_{seen}$, $\lambda_{unseen}$, $\lambda_{lm}$, $\lambda_{cr}$: Weights of losses. $\gamma$: The weight to balance the uncertainty socre.

\textbf{Input}: 

\quad Labeled batch: $\mathcal{X} = \{ (\boldsymbol{x}_i, y_i) \mid i \in \{1, \ldots, B\} \}$. 

\quad Unlabeled batch: $\mathcal{U} = \{ \boldsymbol{u}_i \mid i \in \{1, \ldots, \mu B\} \}$.

\textbf{For ITS}:

\begin{algorithmic}[1] 
\State By $\textbf{Teacher}_{in}$, $\boldsymbol{p}_i^{ITS} = \Phi(f(\text{Aug}_{\text{weak}}(\boldsymbol{u}_i))) \in \mathbb{R}^K$. 
\State By $\textbf{Teacher}_{out}$, $\boldsymbol{p}_i^{OTS} = \psi(f(\text{Aug}_{\text{weak}}(\boldsymbol{u}_i))) \in \mathbb{R}^{K+1}$.

\State $s(\boldsymbol{u}_i) = \gamma \cdot \left(1 - \max(\boldsymbol{p}_i^{ITS})\right) + (1-\gamma) \cdot \boldsymbol{p}_i^{OTS}[-1]$
\Comment{\textcolor{gray}{\textit{Calculate the uncertainty score.}}}
\State By $\textbf{Student}_{in}$, $\boldsymbol{p}_{i}^l = \Phi(f(\text{Aug}_{\text{strong}}(\boldsymbol{x}_i)))$, $\boldsymbol{p}_{i}^u = \psi(f(\text{Aug}_{\text{strong}}(\boldsymbol{u}_i)))$

\State $\mathcal{L}_{ce}^{K}(\mathcal{X}) = \frac{1}{B} \sum_{i=1}^{B} \text{H}(y_i, \boldsymbol{p}_{i}^l)$, $\tilde{y}_i^u = \textit{argmax}(\boldsymbol{p}_i^{ITS})$.

\State $\mathcal{L}_{seen}(\mathcal{U}) = \frac{1}{\mu B} \sum_{i=1}^{\mu B} \text{H}(\tilde{y}_i^u, \boldsymbol{p}_i^u) \cdot \mathbbm{1}(\max(\boldsymbol{p}_i^{ITS}) > \tau) \cdot \mathbbm{1}(\max(\boldsymbol{p}_i^{ITS}) > s(\boldsymbol{u}_i))$
\Comment{\textcolor{gray}{\textit{Select reliable seen-class samples from the unlabeled batch through the $\textbf{Teacher}_{in}$ model and the uncertainty score.}}}

\State $\mathcal{L}_{lm}(\mathcal{U}) = \frac{1}{\mu B} \sum_{i=1}^{\mu B} \mathbb{K}\mathbb{L}(\boldsymbol{p}_{i}^l, \boldsymbol{p}_{i}^{ITS}) \cdot \mathbbm{1}(\max(\boldsymbol{p}_i^{ITS}) > \tau) \cdot \mathbbm{1}(\max(\boldsymbol{p}_i^{ITS}) > s(\boldsymbol{u}_i))$
\Comment{\textcolor{gray}{\textit{Calculate the Logit Matching Loss.}}}
\State ${\mathcal{L}}_{inlier} = {\mathcal{L}}_{ce}^{K} + \lambda_{seen} {\mathcal{L}}_{seen} + \lambda_{lm} {\mathcal{L}}_{lm}$
\Comment{\textcolor{gray}{\textit{Calculate the overall loss for ITS to update its parameters.}}}
\end{algorithmic}

\textbf{For OTS}:

\begin{algorithmic}[1] 
\State By $\textbf{Student}_{out}$, $\boldsymbol{p}_{i}^l = \Phi(f(\text{Aug}_{\text{strong}}(\boldsymbol{x}_i)))$
\State By $\textbf{Student}_{out}$, $\boldsymbol{p}_{sa}= \psi(f(\text{Aug}_{\text{strong}}(\boldsymbol{u}_i)))$, $\boldsymbol{p}_{wa} = \psi(f(\text{Aug}_{\text{weak}}(\boldsymbol{u}_i)))$

\State $\mathcal{L}_{ce}^{K+1}(\mathcal{X}) = \frac{1}{B} \sum_{i=1}^{B} \text{H}(y_i, \boldsymbol{p}_{i}^l)$, $\tilde{y}_i^u = \textit{argmax}(\boldsymbol{p}_i^{OTS})$.

\State $\mathcal{L}_{seen}(\mathcal{U}) = \frac{1}{\mu B} \sum_{i=1}^{\mu B} \text{H}(\tilde{y}_i^u, \boldsymbol{p}_{sa}) \cdot \mathbbm{1}(\max(\boldsymbol{p}_i^{OTS}) > \tau) \cdot \mathbbm{1}(\max(\boldsymbol{p}_i^{OTS}) > s(\boldsymbol{u}_i))$
\State $\mathcal{L}_{unseen}(\mathcal{U}) = \frac{1}{\mu B} \sum_{i=1}^{\mu B} \text{H}(y^{K+1}, \boldsymbol{p}_{sa}) \cdot s(\boldsymbol{u}_i)$
\Comment{\textcolor{gray}{\textit{Generate the ($K$+1)-th class supervision signal to train the model.}}}
\State $\mathcal{L}_{cr}(\mathcal{U}) = \frac{1}{\mu B} \sum_{i=1}^{\mu B} \mathbb{K}\mathbb{L}(\boldsymbol{p}_{wa}, \boldsymbol{p}_{sa})$
\Comment{\textcolor{gray}{\textit{Calculate the consistency regularization loss.}}}
\State $\mathcal{L}_{outlier} = \mathcal{L}_{ce}^{K+1} + \lambda_{seen} \mathcal{L}_{seen} + \lambda_{unseen} \mathcal{L}_{unseen} + \lambda_{cr} \mathcal{L}_{cr}$
\Comment{\textcolor{gray}{\textit{Calculate the overall loss for OTS to update its parameters.}}}
\end{algorithmic}

\textbf{Summary}: During the training process, ITS and OTS collaborate to generate the uncertainty score, which aids the model in detecting and filtering unseen-class samples, thereby further enhancing the model's performance.
\end{algorithm*}

\subsection{Overall Framework}
In this section, we detail the DTS training process, elucidating how both Inlier Teacher-Student (ITS) and  Outlier Teacher-Student (OTS) harmoniously optimize their objectives. Essentially, beyond solely relying on labeled data for generating supervisory signals during the teacher model's pre-training phase, both ITS and OTS utilize both labeled data $L$ and unlabeled data $U$ throughout their training. We have tailored specific loss functions for various tasks to ensure that the holistic optimization goals of Diverse Teacher-Students remain aligned during the training.

The cumulative optimization objective for ITS is:
\begin{equation}
\begin{aligned}
{\mathcal L}_{inlier}={\mathcal L}_{ce}^{K}(L)+\lambda_{seen}{\mathcal L}_{seen}(U)+\lambda_{lm}{\mathcal L}_{lm}(U)\,,
\end{aligned}
\tag{9}
\end{equation}
where $\lambda_{seen}$ and $\lambda_{lm}$ are the weights assigned to each learning objective. Throughout the training process, the ITS leverages both labeled data and vetted samples from unlabeled data to generate supervisory signals, thereby enhancing the classification prowess for seen classes. Concurrently, we employ a logit matching approach to closely align the student model's predictions with those of the teacher model, thereby further enhancing performance.

The comprehensive optimization objective for OTS is:
\begin{equation}
\begin{aligned}
\mathcal{L}_{outlier} &= \mathcal{L}_{ce}^{K+1}(L) + \lambda_{seen}\mathcal{L}_{seen}(U) \\
& \quad + \lambda_{unseen}\mathcal{L}_{unseen}(U) + \lambda_{cr}\mathcal{L}_{cr}(U),
\end{aligned}
\tag{10}
\end{equation}
where, $\lambda_{seen}$, $\lambda_{unseen}$, and $\lambda_{cr}$ denote the weights for each learning objective. 

Mirroring the ITS, the OTS also trains using both labeled data and carefully curated samples from the unlabeled dataset. However, the OTS places a heightened focus on unseen-class detection. To bolster the detection of unseen classes, we employ an uncertainty score to subtly soft separation unseen-class samples from unlabeled data, integrating the ($K$+1)-th class supervisory signal into the training. This approach not only significantly enhances the detection capabilities for unseen classes but also, in turn, boosts the effectiveness of the uncertainty score, further optimizing the model. Besides, to further improve the robustness of our model, we impose a consistent regularization loss $\mathcal{L}_{cr}$ on the OTS.

Within the DTS framework, both ITS and OTS are intertwined and influence each other through the Soft-weighting module. Collectively, they optimize both the classification of seen classes and the detection of unseen classes. In Algorithm \ref{alg:algorithm2}, we illustrate how ITS and OTS collaborate by employing the Soft-weighting module to optimize the overall framework.

\begin{algorithm*}[t]
\footnotesize
\caption{Inference of \textbf{DTS}}
\label{alg:algorithm3}
\textbf{Setup}:

\quad $f(\cdot)$: Feature extractor. $\Phi$: The K-classifier. $\psi$: The ($K$+1)-classifier. $\gamma$: The weight to balance the uncertainty socre.

\quad $\textbf{Student}_{in} : f(\cdot) + \Phi$

\quad $\textbf{Student}_{out} : f(\cdot) + \psi$

\textbf{Input}: 

\quad Test batch: $\textit{T} = \{ (\boldsymbol{x}_i, y_i) \mid i \in \{1, \ldots, B\} \}$.

\quad Unlabeled data: $U=\{\boldsymbol {u}_1,\boldsymbol {u}_2,...,\boldsymbol {u}_n\}$.

\quad \textbf{For test.}

\begin{algorithmic}[1] 
\State By $\textbf{Student}_{in}$, $\boldsymbol{p}_i = \Phi(f(\boldsymbol{x}_i))$
\Comment{\textcolor{gray}{\textit{Get classification predictions of K classes}}}
\State By $\textbf{Student}_{out}$, $\boldsymbol{p}_i^{\prime} = \psi(f(\boldsymbol{x}_i))$
\Comment{\textcolor{gray}{\textit{Get classification predictions of K+1 classes}}}
\State $y_i^p=\textit{argmax}(\boldsymbol{p}_i)$
\Comment{\textcolor{gray}{\textit{For each test sample, get its prediction label.}}}
\end{algorithmic}

\quad \textbf{Calculate the AUROC of DTS in unlabeled data.}
\Comment{\textcolor{gray}{\textit{Ability to detect unseen classes.}}}

\begin{algorithmic}[1] 
\State By $\textbf{Student}_{in}$, $\boldsymbol{p}_i = \Phi(f(\boldsymbol{u}_i))$
\Comment{\textcolor{gray}{\textit{Get classification predictions of K classes}}}
\State By $\textbf{Student}_{out}$, $\boldsymbol{p}_i^{\prime} = \psi(f(\boldsymbol{u}_i))$
\Comment{\textcolor{gray}{\textit{Get classification predictions of K+1 classes}}}

\State $s(\boldsymbol{u}_i) = \gamma \cdot \left(1 - \max(\boldsymbol{p}_i)\right) + (1-\gamma) \cdot \boldsymbol{p}_i^{\prime}[-1]$
\Comment{\textcolor{gray}{\textit{Calculate the uncertainty score.}}}

\end{algorithmic}
\textbf{Inference}: 

\quad Calculate accuracy with the true label $y_i$ and predicted label $y_i^p$.

\quad Calculate AUROC with the true label $y_i$ and the uncertainty score $s(\boldsymbol{u}_i)$.
\end{algorithm*}

\subsection{Inference}
As shown in algorithm \ref{alg:algorithm3}, we use the $\textbf{Student}_{in}$ model in ITS to classify test dataset with only seen classes, and use the uncertainty score to distinguish the ID samples and OOD samples in unlabeled data.

\section{Experiments}

\subsection{Setup}
\textbf{Datasets.} We conducted a comprehensive evaluation of the DTS framework using well-known public datasets, including SVHN \cite{netzer2011reading}, CIFAR-10, CIFAR-100 \cite{krizhevsky2009learning} and STL-10 \cite{DBLP:journals/jmlr/CoatesNL11}. To encompass a broad range of safe semi-supervised learning scenarios, our experiments were designed with varying class mismatch ratios and diverse sizes of labeled datasets. This approach ensures a thorough assessment of the DTS model's performance across different settings and conditions.

\textbf{Baselines.}
Our evaluation of the DTS model involves a comparative analysis against various baseline models, focusing on a test set comprised solely of seen-class data. For standard SSL benchmarks, DTS is compared against established models like the \textbf{Pi-Model} \cite{DBLP:conf/nips/SajjadiJT16}, \textbf{Pseudo-Labeling} \cite{lee2013pseudo}, Virtual Adversarial Training (\textbf{VAT}) \cite{DBLP:journals/pami/MiyatoMKI19}, \textbf{FixMatch} \cite{DBLP:conf/nips/SohnBCZZRCKL20}, \textbf{FlexMatch} \cite{zhang2021flexmatch} and \textbf{FreeMatch} \cite{wang2022freematch}. For safe SSL benchmarks, we consider a range of published works as baselines, including \textbf{UASD} \cite{DBLP:conf/aaai/ChenZLG20}, \textbf{MTCF} \cite{DBLP:conf/eccv/YuIIA20}, \textbf{CL} \cite{DBLP:conf/aaai/Cascante-Bonilla21}, \textbf{TOOR} \cite{DBLP:journals/tmm/Huang0023}, \textbf{Safe-Student} \cite{DBLP:conf/cvpr/HeHLY22} and \textbf{IOMatch} \cite{DBLP:conf/iccv/00100S023}. Additionally, we include a supervised learning method, trained with all available labeled data, as an additional baseline for comprehensive comparison.

\textbf{Evaluation Metrics.}
In our evaluation framework, we adopt accuracy and AUROC as pivotal performance metrics. Accuracy is the primary measure in semi-supervised learning contexts, particularly crucial in tasks like image classification \cite{DBLP:journals/corr/SimonyanZ14a}. AUROC emerges as a secondary yet significant metric, reflecting the ability to discern between ID and OOD samples in unlabeled data, with 50\% as the the random-guessing baseline \cite{DBLP:journals/corr/abs-2306-09301}. An higher AUROC can make better use of unlabeled data in safe SSL, thereby aiding in the enhancement of accuracy. Our proposed DTS model is designed to pursue high accuracy while ensuring the model's capability to differentiate between ID and OOD samples, thereby amplifying its potential under safe SSL scenario.

\textbf{Fairness of Comparisons.} To ensure fairness, all baselines utilize the same neural network as the training backbone, namely, WRN-28-2 \cite{DBLP:conf/bmvc/ZagoruykoK16}. Additionally, we have removed the EMA mechanism from all baselines to simplify and enhance the clarity of the experiments.

\subsection{Main Results}
\subsubsection{SVHN, CIFAR-10, and CIFAR-100}
The SVHN dataset bears resemblance to CIFAR-10 in that both contain ten distinct classes. For the sake of simplicity, we align the six animal classes in CIFAR-10 and the first six digital classes as seen classes. The CIFAR-100 dataset is more complex, consisting of 20 super-classes, each composed of 5 common classes. Also for simplicity, we designate the first 50 classes of CIFAR-100 as seen classes, while the remaining classes are considered unseen classes.

Specifically, for the SVHN and CIFAR-10 datasets, we set the size of the labeled set at 2,400 and the unlabeled set at 20,000. For the CIFAR-100 dataset, the labeled set is sized at 5,000 and the unlabeled set remains at 20,000. We train the network using Stochastic Gradient Descent (SGD) with a learning rate of 0.128, similar to the approach in \cite{DBLP:conf/cvpr/XieLHL20}, a momentum of 0.9, and a weight decay of \(5e^{-4}\). Except that the batch size on STL-10 dataset is 128, we apply a consistent set of hyperparameters across all tasks, specifically: \{$\lambda_{seen}=\lambda_{lm}=0.25$, $\lambda_{unseen}=0.1$, $\lambda_{cr}=0.3$, $\mu=7$, $\tau=0.85$, $n=256$, $\gamma=0.5$, $N_e=400$, $N_i=3$\}. Here, `n' denotes the labeled batch size, $N_e$ the number of epochs, and $N_i$ the number of update iterations within the teacher-student framework. For instance, for the SVHN dataset, training involves three iterations of optimization, each encompassing 400 epochs with a labeled batch size of 256.

Table \ref{table of results} and table \ref{table of results2} present the mean accuracy across five runs, accounting for varying class mismatch ratios within the SVHN, CIFAR-10, CIFAR-100 and STL-10 datasets. Notably, the remarkable performance of FlexMatch \cite{zhang2021flexmatch} and FreeMatch \cite{wang2022freematch} on the CIFAR-10 dataset underscores the efficacy of contemporary semi-supervised learning approaches in addressing simple safe SSL scenarios. However, their performance on CIFAR-100 reveals limitations when confronted with more intricate safe SSL scenarios. Consequently, considering the performance across all datasets, our proposed DTS demonstrates superior overall performance.

For example, with a 30\% class mismatch ratio on CIFAR-100, our approach attains a mean accuracy of 71.4\%, which is roughly 1.8\% higher than FreeMatch \cite{wang2022freematch} and an improvement of approximately 2.7\% over the IOMatch \cite{DBLP:conf/iccv/00100S023}. Similarly, at a mismatch ratio of 0.6 on CIFAR-100, DTS achieves 70.0\% accuracy, outperforming FreeMatch \cite{wang2022freematch} by 4.7\% and Safe-student \cite{DBLP:conf/cvpr/HeHLY22} by 1.8\%. These results underscore the superior performance of DTS in complex safe SSL scenarios.

\begin{table*}[t]
    \setlength{\abovecaptionskip}{0.cm}
    \captionsetup{font={footnotesize}}
	\caption{Seen-class classification accuracy (\%) of different methods on SVHN and CIFAR-10.}
	\label{tab:comparison}
	\renewcommand{\arraystretch}{1}
        \renewcommand{\tabcolsep}{12pt}
    \centering
    \scalebox{0.9}{
\begin{tabular}{l|cc|cc}
    \toprule
                                  & \multicolumn{2}{c|}{SVHN}                                                               & \multicolumn{2}{c}{CIFAR-10}                                       
                                  \\

\multirow{-2}{*}{Method/Datasets} & \begin{tabular}[c]{@{}c@{}}ratio= 0.3\end{tabular} & ratio=0.6                       & \begin{tabular}[c]{@{}c@{}}ratio= 0.3\end{tabular} & ratio=0.6 \\
\midrule
Supervised                        & 92.1±0.1                                             & 92.1±0.1                        & 76.3±0.4                                             & 76.3±0.4  \\
\midrule
Pi-Model \cite{DBLP:conf/nips/SajjadiJT16}                          & 93.9±0.6                                             & 93.8±0.8                        & 75.7±0.7                                             & 74.5±1.0  
 \\
PL \cite{lee2013pseudo}                                & 93.2±0.6                                             & 92.7±0.7                        & 75.8±0.8                                             & 74.6±0.7  
 \\
VAT \cite{DBLP:journals/pami/MiyatoMKI19}                               & 94.6±0.5                                             & 93.3±0.5                        & 76.9±0.6                                             & 75.0±0.5  
 \\
FixMatch \cite{DBLP:conf/nips/SohnBCZZRCKL20}                          & 94.6±0.6                                             & 94.5±0.6                        & 81.5±0.2                                             & 80.9±0.3  
 \\
 FlexMatch \cite{zhang2021flexmatch}                          & 95.6±0.5                                             & 95.0±0.2                        & 88.4±0.4                                             & 87.6±0.6  
 \\
  FreeMatch \cite{wang2022freematch}                          & 95.9±0.5                                             & 95.2±0.2                        & \textbf{89.3±0.7}                                             & \textbf{88.5±0.5}  
 \\
\midrule
UASD \cite{DBLP:conf/aaai/ChenZLG20}                              & 93.1±0.3                                             & 92.5±0.4                        & 77.6±0.4                                             & 76.0±0.4  
 \\
MTCF \cite{DBLP:conf/eccv/YuIIA20}                               & 96.2±0.5                                             & 95.7±0.5                        & 85.5±0.6                                             & 81.7±0.5  
 \\
CL \cite{DBLP:conf/aaai/Cascante-Bonilla21}                                & 95.4±0.4                                             & 95.1±0.4                        & 83.2±0.4                                             & 82.1±0.4  
 \\

TOOR \cite{DBLP:journals/tmm/Huang0023}  &  95.1±0.2  &  94.7±0.2  &  78.9±0.6  &  78.1±0.5   \\
Safe-Student\cite{DBLP:conf/cvpr/HeHLY22}                      & 94.8±0.3                                             & 94.1±0.3                        & 85.7±0.3                                             & 83.8±0.1  
 \\
 IOMatch\cite{DBLP:conf/iccv/00100S023}                      & 96.1±0.2                                             & 95.7±0.1                        & 89.1±0.4                                             & 88.0±0.5 
 \\
\midrule
\textbf{DTS}                               & \textbf{96.5±0.2}                      & \textbf{96.1±0.1} & 87.7±0.5                                             & 86.4±0.4  
 \\
\bottomrule
\end{tabular}
    }
    \label{table of results}
\end{table*}

\begin{table*}[t]
    \setlength{\abovecaptionskip}{0.cm}
    \captionsetup{font={footnotesize}}
	\caption{Seen-class classification accuracy (\%) of different methods on CIFAR-100 and STL-10.}
	\label{tab:comparison2}
	\renewcommand{\arraystretch}{1}
        \renewcommand{\tabcolsep}{12pt}
    \centering
    \scalebox{0.9}{
\begin{tabular}{l|cc|c}
    \toprule
                                                                      & \multicolumn{2}{c|}{CIFAR-100}     
                                  &
                                  \multicolumn{1}{c}{STL-10}   
                                  \\

\multirow{-2}{*}{Method/Datasets}  & \begin{tabular}[c]{@{}c@{}}ratio= 0.3\end{tabular} & ratio=0.6 & ratio=none\\
\midrule
Supervised                          & 58.6±0.5                                             & 58.6±0.5 & 63.4±1.0 \\
\midrule
Pi-Model \cite{DBLP:conf/nips/SajjadiJT16}                            & 59.4±0.3                                             & 57.9±0.3 & 64.5±0.9
 \\
PL \cite{lee2013pseudo}                                  & 60.2±0.3                                             & 57.5±0.6  & 64.7±0.9
 \\
VAT \cite{DBLP:journals/pami/MiyatoMKI19}                                 & 61.8±0.4                                             & 59.6±0.6  & 63.9±0.4
 \\
FixMatch \cite{DBLP:conf/nips/SohnBCZZRCKL20}                            & 65.9±0.3                                             & 65.2±0.3 & 67.1±0.5
 \\
 FlexMatch \cite{zhang2021flexmatch}      & 68.7±0.3    & 65.2±0.1 & 85.8±0.6
 \\
FreeMatch \cite{wang2022freematch}      & 69.6±0.1    & 65.3±0.2 & 85.9±0.5
 \\
\midrule
UASD \cite{DBLP:conf/aaai/ChenZLG20}                                & 61.8±0.4                                             & 58.4±0.5 & 62.8±0.2
 \\
MTCF \cite{DBLP:conf/eccv/YuIIA20}                                & 63.1±0.6                                             & 61.1±0.3 & 66.5±0.5
 \\
CL \cite{DBLP:conf/aaai/Cascante-Bonilla21}                                  & 63.6±0.4                                             & 61.5±0.5 & 67.1±1.0
 \\

TOOR \cite{DBLP:journals/tmm/Huang0023}    &  64.4±0.5  &63.6±0.6 & 78.0±1.0 \\
Safe-Student\cite{DBLP:conf/cvpr/HeHLY22}                        & 68.4±0.2                                             & 68.2±0.1 & 84.6±0.7
 \\
 IOMatch\cite{DBLP:conf/iccv/00100S023}                       & 68.7±0.6                                             & 65.6±0.1 & 85.6±0.5
 \\
\midrule
\textbf{DTS}                                 & \textbf{71.4±0.3}                                             & \textbf{70.0±0.5} & \textbf{87.9±0.3}
 \\
\bottomrule
\end{tabular}
    }
    \label{table of results2}
\end{table*}

\subsubsection{STL-10}
To better align with real-world application scenarios, we conducted additional experiments on the STL-10 dataset, as illustrated in Table \ref{table of results2}. Similar to the CIFAR-10 dataset, we categorized animal classes as seen classes and the remaining categories as unseen classes.

For these experiments, we utilized the entire STL-10 dataset, which consists of 5,000 labeled samples and 100,000 unlabeled images, thereby setting the class mismatch ratio to none. This approach more closely mirrors real-world application conditions, where prior information about the unlabeled set is often unavailable.

In summary, as shown in the last column of Table \ref{table of results2}, some existing methods exhibit rather average performance on the STL-10 dataset. This is attributed to the dataset's large volume of unlabeled samples compared to a mere 5,000 labeled images, challenging models to effectively discern between seen and unseen classes within the unlabeled set and thus hindering the acquisition of valuable information, leading to suboptimal training outcomes. Conversely, our proposed DTS method can perform soft-separation of seen and unseen classes within the unlabeled set without requiring additional prior information, enabling the extraction of a substantial amount of valuable information and achieving optimal performance.

\begin{figure}[t]
  \centering
  \begin{subfigure}[b]{0.45\columnwidth}
    \centering
    \includegraphics[width=\textwidth]{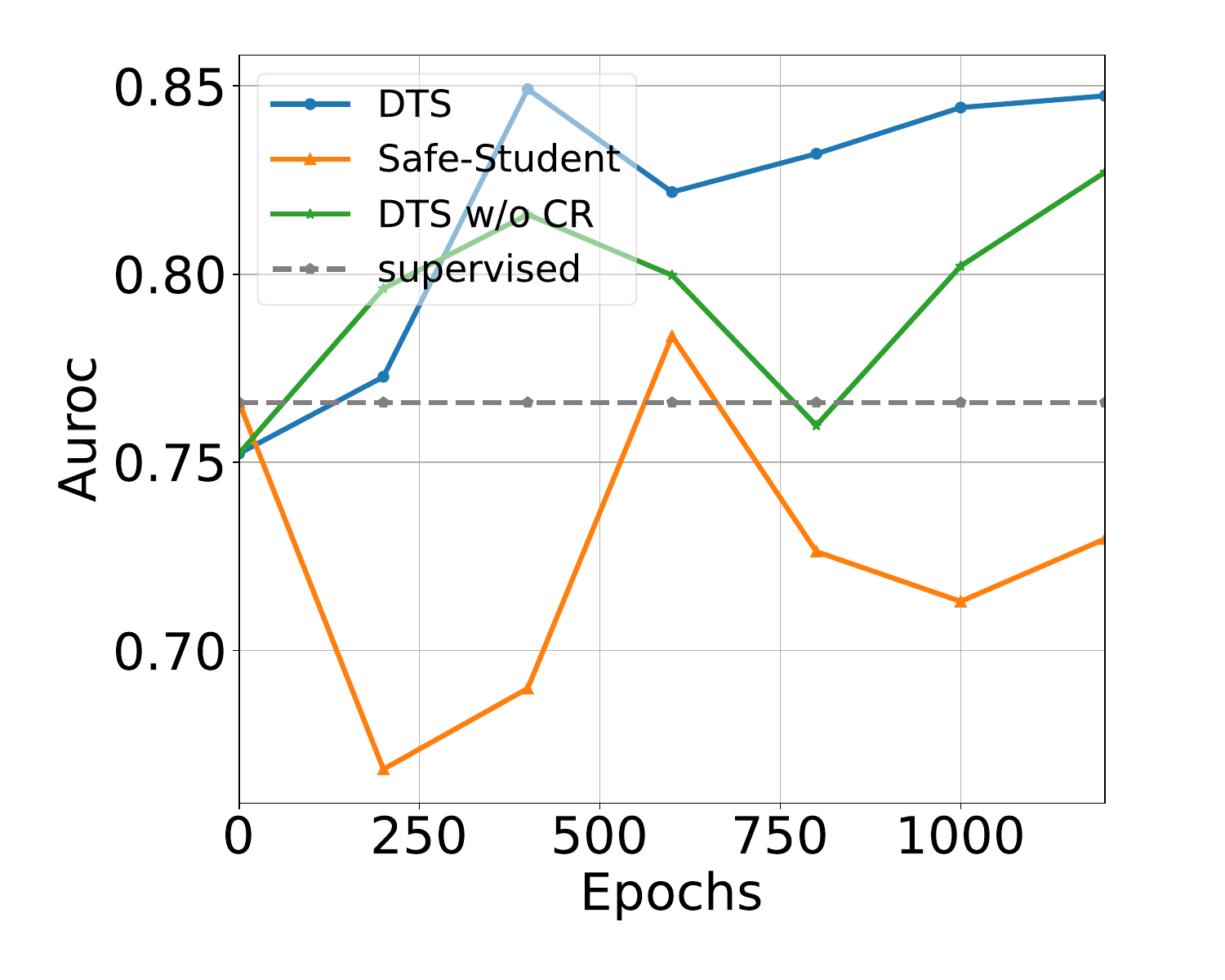}
    \caption{30\% Class Mismatch Ratio}
    \label{fig:30}
  \end{subfigure}
  \hfill 
  \begin{subfigure}[b]{0.45\columnwidth}
    \centering
    \includegraphics[width=\textwidth]{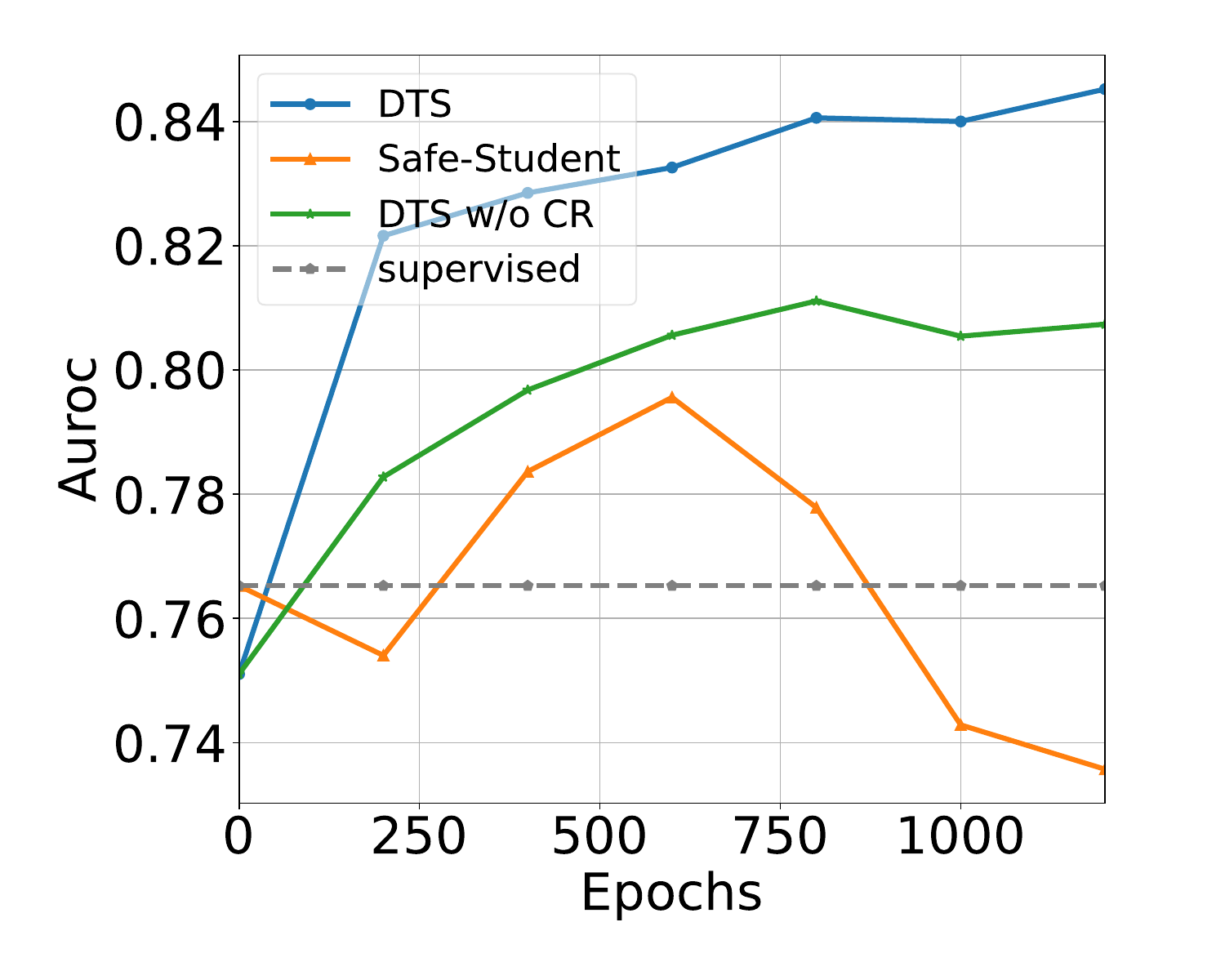}
    \caption{60\% Class Mismatch Ratio}
    \label{fig:60}
  \end{subfigure}
  \caption{(a) and (b) respectively depict the AUROC trend for our proposed DTS and SAFE-STUDENT models, as well as the supervised baseline, under varying Labeled/Unlabeled Classes Mismatch ratios on CIFAR-10.}
  \label{fig of auroc}
\end{figure}

\subsection{Unseen-Class Detection}
A key innovation of our proposed DTS model lies in its dual teacher-student framework, which separates and concurrently optimizes the tasks of seen-class classification and unseen-class detection. This approach effectively enhances the model's ability to distinguish between seen and unseen-classes during the training process. As demonstrated in parts Figure \ref{fig:30} and Figure \ref{fig:60}, DTS maintains and incrementally enhances the stability of the AUROC metric. In contrast, Safe-Student \cite{DBLP:conf/cvpr/HeHLY22}, which relies on a single model for dual tasks, exhibits considerable AUROC volatility throughout training, and the improvement of accuracy is at the expense of AUROC. Notably, DTS achieves an approximate 10\% increase in AUROC at both 30\% and 60\% mismatch ratios on the CIFAR-10 dataset.

Thus, we can conclude that the capacity to discriminate between seen and unseen classes can be markedly enhanced by training the two tasks both separately and collaboratively. This strategy optimizes the use of unlabeled data, further releasing the potential for model optimization.

\subsection{Ablation Analysis}
\begin{figure*}[t]
\centering
    \subfloat[m=600]
    {
        \label{fig:600}\includegraphics[width=0.32\textwidth]{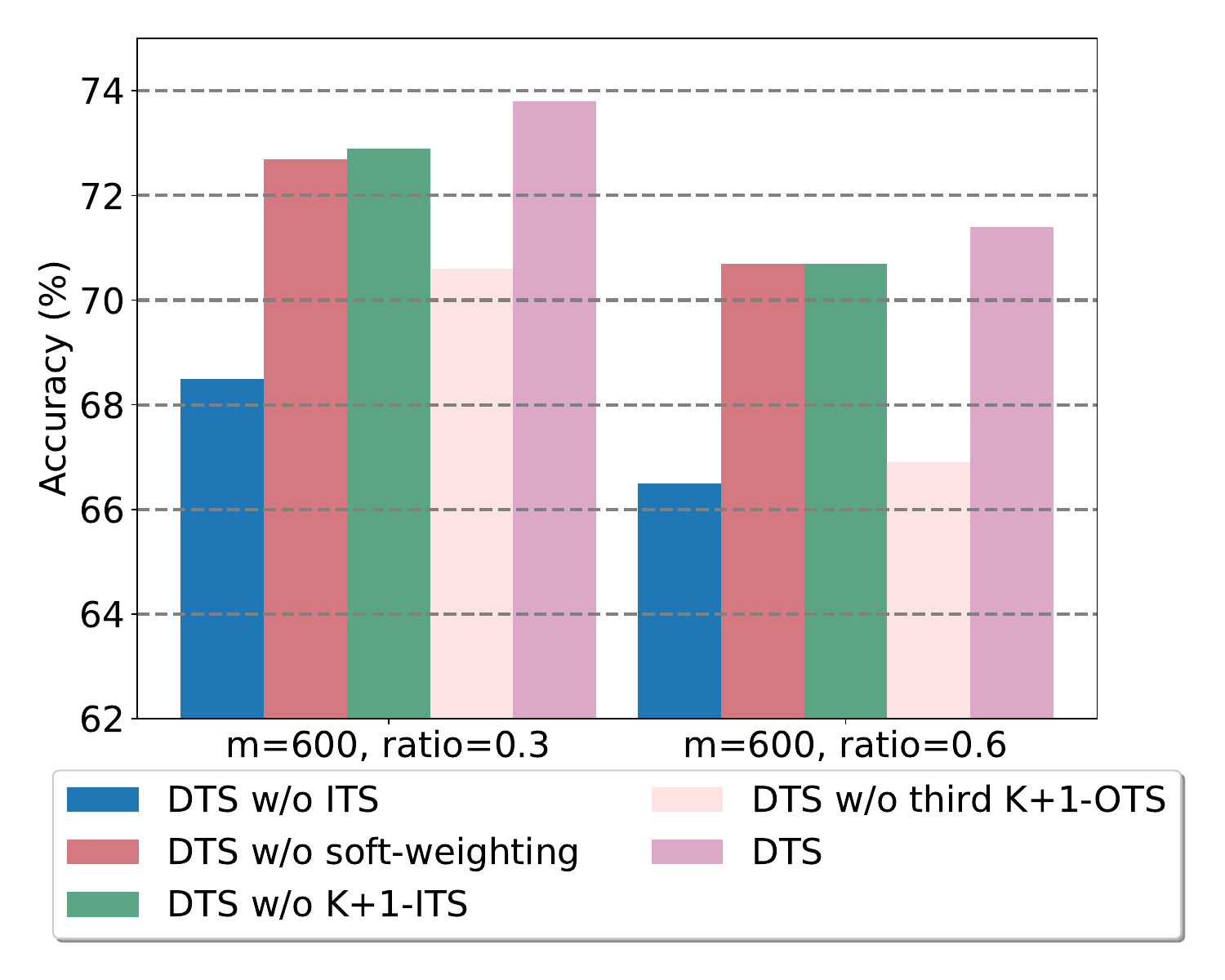}
    }
    \subfloat[m=1200]
    {
        \label{fig:1200}\includegraphics[width=0.32\textwidth]{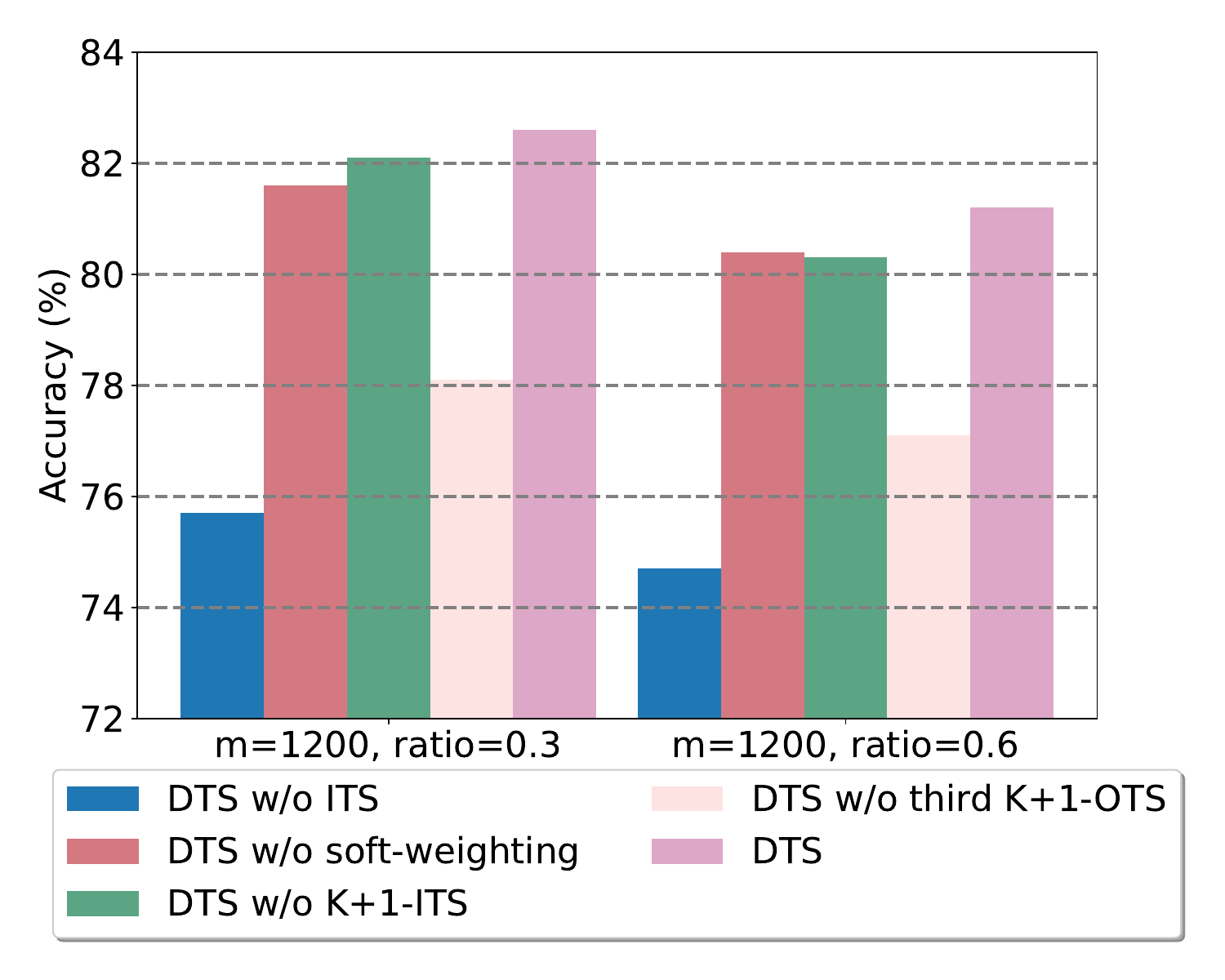}
    }
    \subfloat[m=2400]
    {
        \label{fig:2400}\includegraphics[width=0.32\textwidth]{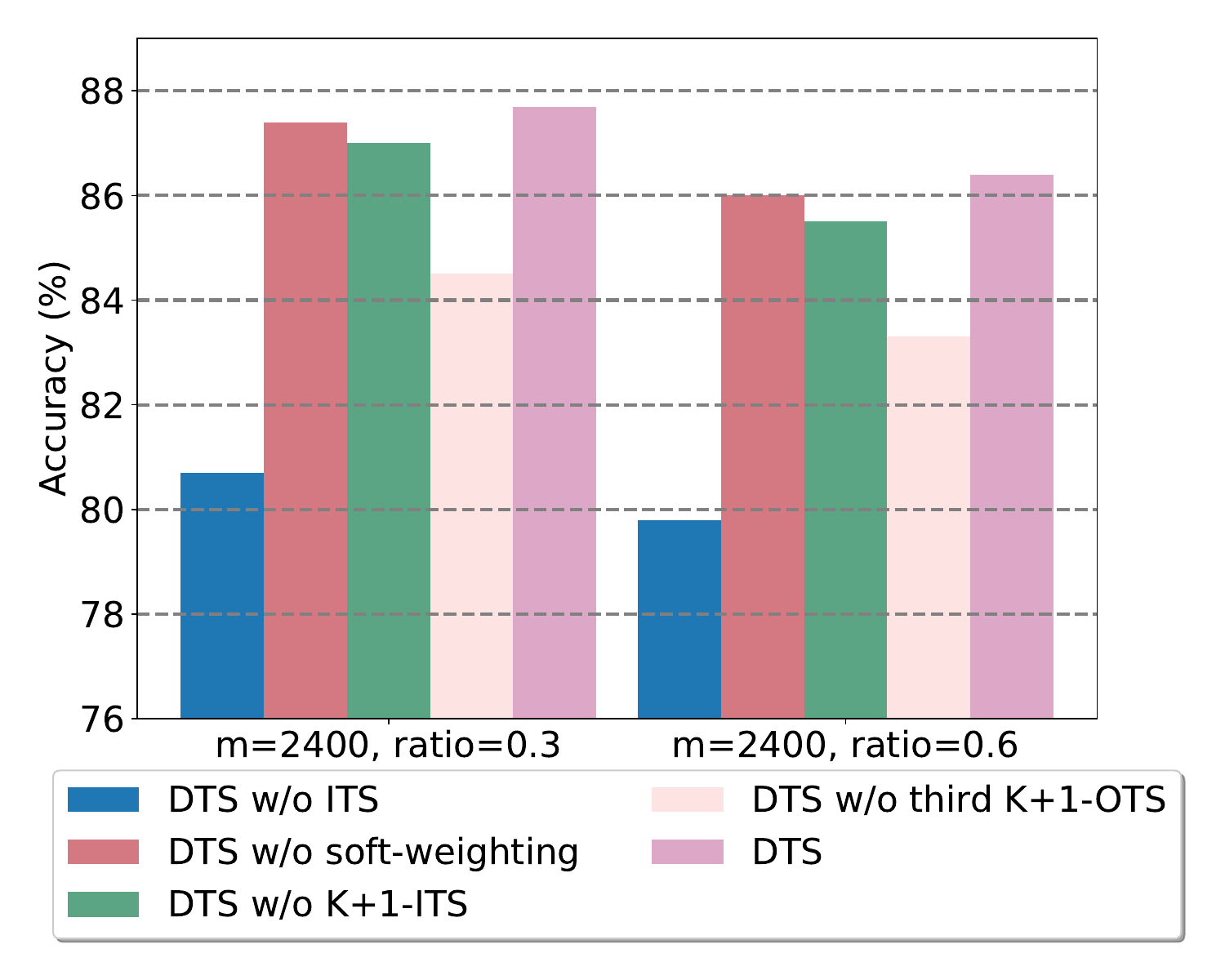}
    }
\caption{Ablation Analysis: (a), (b), and (c) depict the impact of various modules on the classification performance of the DTS framework across different values of `m' and class mismatch ratios, illustrating the contribution of each component under varying conditions.}
\label{fig of abla}
\end{figure*}

To validate the effectiveness of the modules we proposed, we conducted extensive ablation studies on the respective modules. Specifically, Figure \ref{fig of abla} presents the results of these ablation studies on CIFAR-10 with varying settings. It encompasses various versions of the DTS framework, which include:

\textbf{DTS w/o ITS}: DTS without the ITS module, where instead of using two models to tackle two key tasks independently, a single model OTS with a ($K$+1)-classifier is utilized for the classification of seen classes and detection of unseen classes.

\textbf{DTS w/o Soft-weighting}: DTS without the Soft-weighting module, meaning it does not employ the uncertainty score to softly separate unseen-class samples within the unlabeled data. Rather, it sets a threshold (0.85) for the uncertainty score to select reliable unseen-class samples for training in the (K+1)-th class setting.

\textbf{DTS w/o K+1-ITS} and \textbf{DTS w/o K+1-OTS}: DTS without utilizing the (K+1)-th class samples for generating supervisory signals in training, in which case, DTS is separated into two independent models, ITS and OTS with K-classifier. Specifically, the ITS follows traditional safe SSL approaches, disregarding unseen-class samples and selecting reliable seen-class samples from unlabeled data based on a confidence threshold (0.85); whereas OTS, similar to ITS, but still contemplates the detection of unseen classes, using $1-\max(\boldsymbol{p}_i)$ as the uncertainty score. It then selects samples based on a threshold (0.5) to promote a distribution that approaches uniformity.

Specifically, ablation studies were conducted on the CIFAR-10 dataset with varying sizes of labeled data, namely m=600, m=1200, and m=2400. Additionally, each of these sizes corresponded to different class mismatch ratios, 0.3 and 0.6. The results of these experiments, as illustrated in Figure \ref{fig of abla}, strongly substantiate the efficacy of the individual components within the DTS framework. It is noteworthy that although some results for DTS without the (K+1)-th class (DTS w/o K+1) are slightly worse than the standard DTS, the corresponding AUROC displays considerable instability, closely paralleling the AUROC of Safe-Student \cite{DBLP:conf/cvpr/HeHLY22} as depicted in Figure \ref{fig of auroc}. This further underscores the enhanced performance and stability offered by the standard DTS framework.

\begin{figure}[t]
\centering
    \subfloat[30\% Class MisMatch Ratio]
    {
        \label{fig:subfig3}\includegraphics[width=0.45\columnwidth]{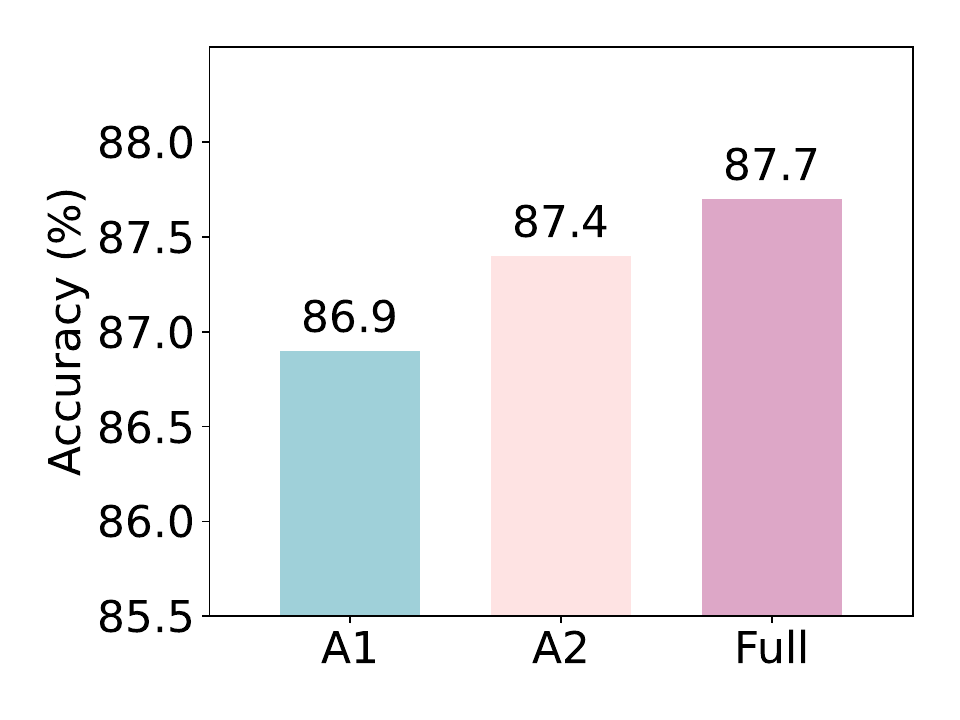}
    }
    \subfloat[60\% Class MisMatch Ratio]
    {
        \label{fig:subfig4}\includegraphics[width=0.45\columnwidth]{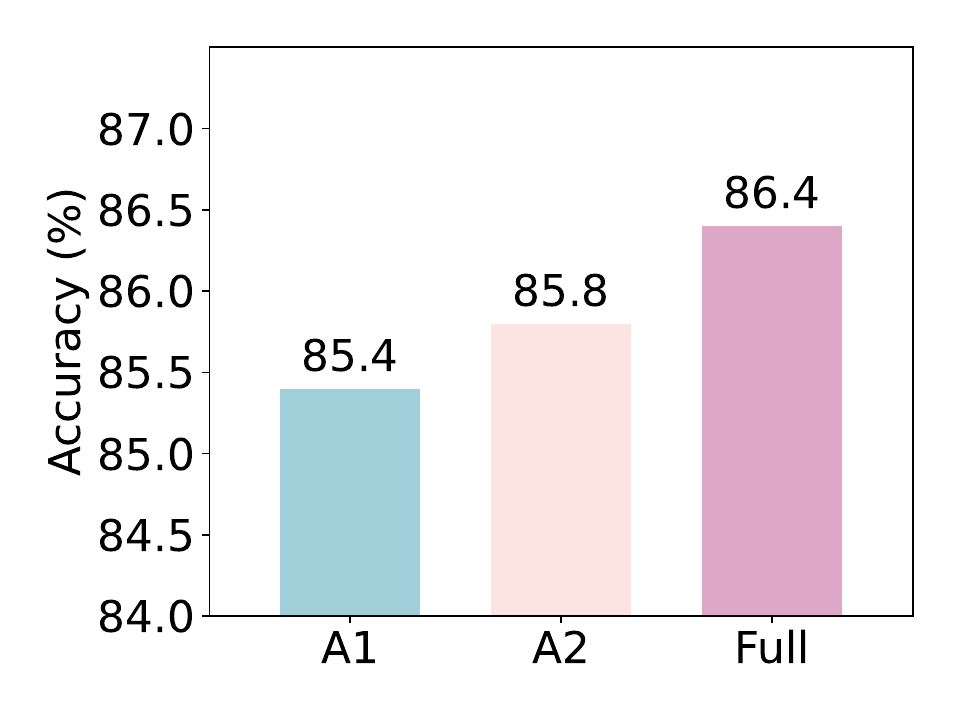}
    }
\caption{Ablation Study on Logit Matching and Consistency Regularization. `A1' signifies DTS without Logit Matching, `A2' indicates DTS lacking Consistency Regularization, and `Full' refers to the complete DTS framework with all components integrated.}
\label{fig:abla-cr and da}
\end{figure}

Additionally, we have investigated the impact of Logit Matching and Consistency Regularization on the DTS framework. As depicted in Figure \ref{fig:abla-cr and da}, both factors influence the classification performance of DTS; however, Logit Matching has a more pronounced effect on enhancing DTS's performance. This is because it enables the teacher model to provide strong guidance to the student model throughout the training process. Consistency Regularization, on the other hand, tends to bolster the model's robustness, resulting in a smoother variation in AUROC.

As shown in Figure \ref{fig of auroc}, although the AUROC of DTS without Consistency Regularization (DTS w/o CR) remains stable and gradually improves, it does not achieve the same level of steadiness as the full DTS, exhibiting relatively greater fluctuations.

\begin{figure*}[t]
\setlength{\abovecaptionskip}{0.cm}
  \centering
  \begin{subfigure}{0.32\textwidth}
    \includegraphics[width=1\linewidth]{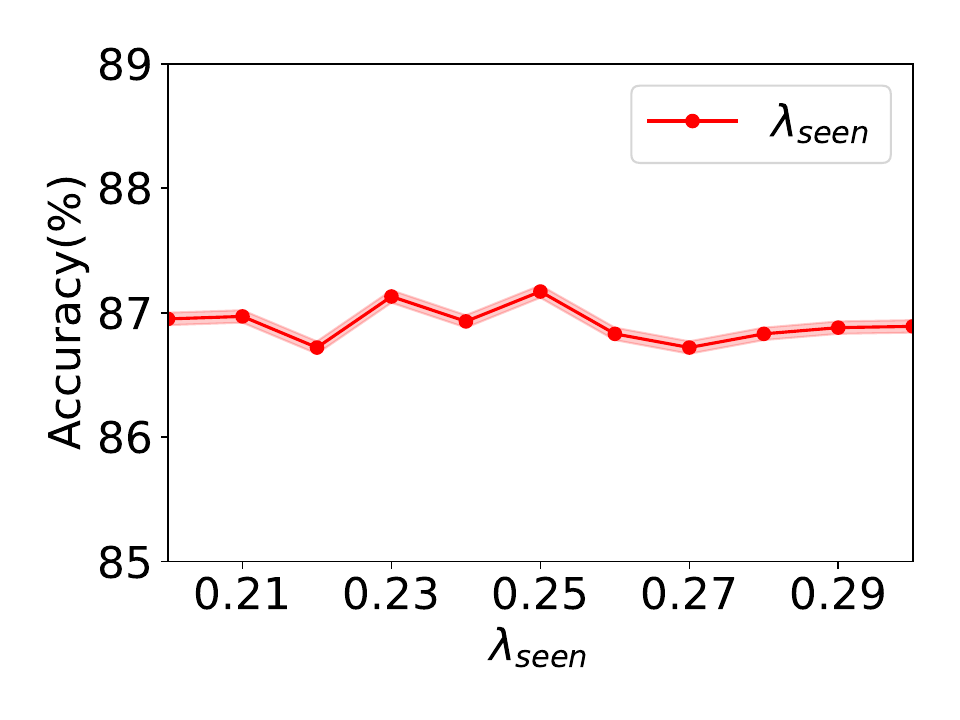}
    \label{ctk}
  \end{subfigure}
  \hfill
  \begin{subfigure}{0.32\linewidth}
    \includegraphics[width=1\textwidth]{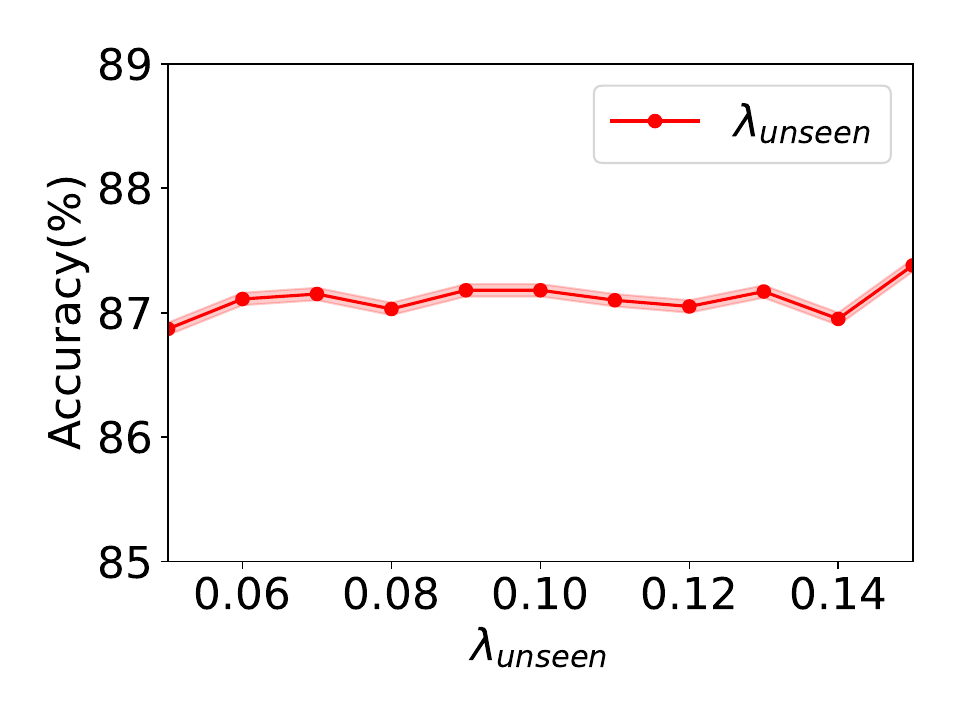}
    \label{ctu}
  \end{subfigure}
  \hfill
  \begin{subfigure}{0.32\linewidth}
    \includegraphics[width=1\textwidth]{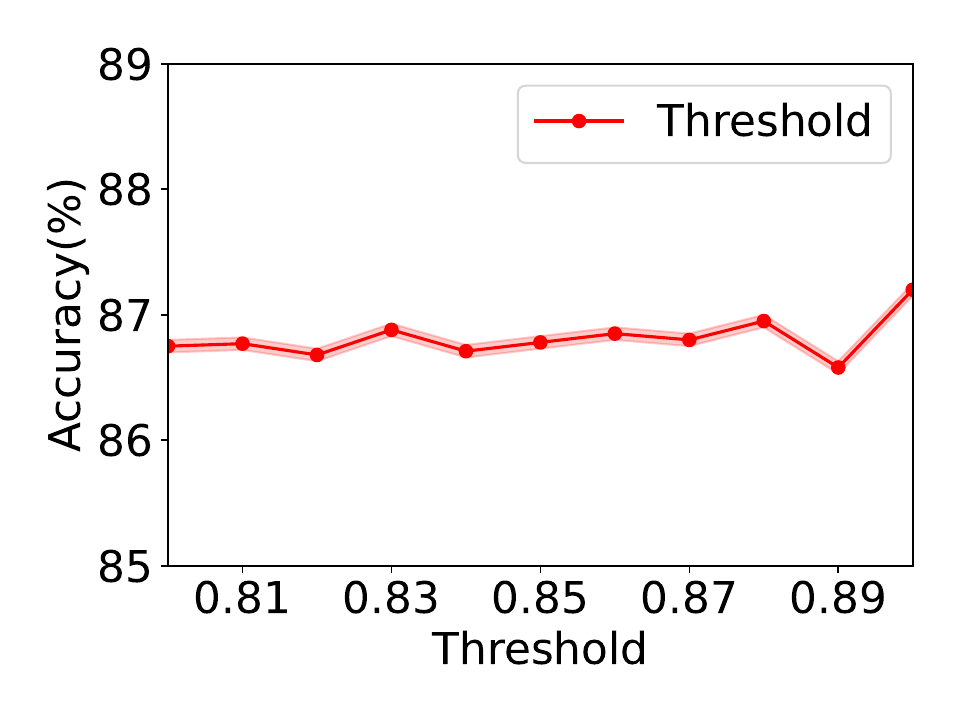}
    \label{ctu2}
  \end{subfigure}
\captionsetup{font={footnotesize}}
\caption{Hyperparameter Impact Analysis: This graph illustrates the influence of three parameters—\( \lambda_{seen} \), \( \lambda_{unseen} \), and \( \text{Threshold} \)—on the accuracy performance of our proposed DTS framework when applied to the CIFAR-10 dataset under a class mismatch ratio of 0.4, revealing the trend of how each hyperparameter affects the model's effectiveness.}
\label{fig of hyperpara}
\end{figure*}

\subsection{Sensitivity of Hyperparameters}
Figure \ref{fig of hyperpara} explores the sensitivity of key hyperparameters, including the loss function weights ($\lambda_{seen}$ and $\lambda_{unseen}$) and the confidence threshold. These parameters play pivotal roles in leveraging unlabeled data. Specifically, in our experiments, we set $\lambda_{seen}$ at 0.25 and $\lambda_{unseen}$ at 0.10, with a confidence threshold of 0.85. As Figure \ref{fig of hyperpara} illustrates, under the CIFAR-10 dataset with a class mismatch ratio of 0.4, even within a ±0.1 deviation from these values, neither $\lambda_{seen}$, $\lambda_{unseen}$, nor the threshold significantly affect the classification performance of DTS. This highlights the robustness of our DTS framework against fluctuations in these hyperparameters.

\begin{figure*}[t]
\centering
    \subfloat[train loss on CIFAR-10]
    {
        \label{fig:  train_loss_cifar10_0.3}\includegraphics[width=0.32\textwidth]{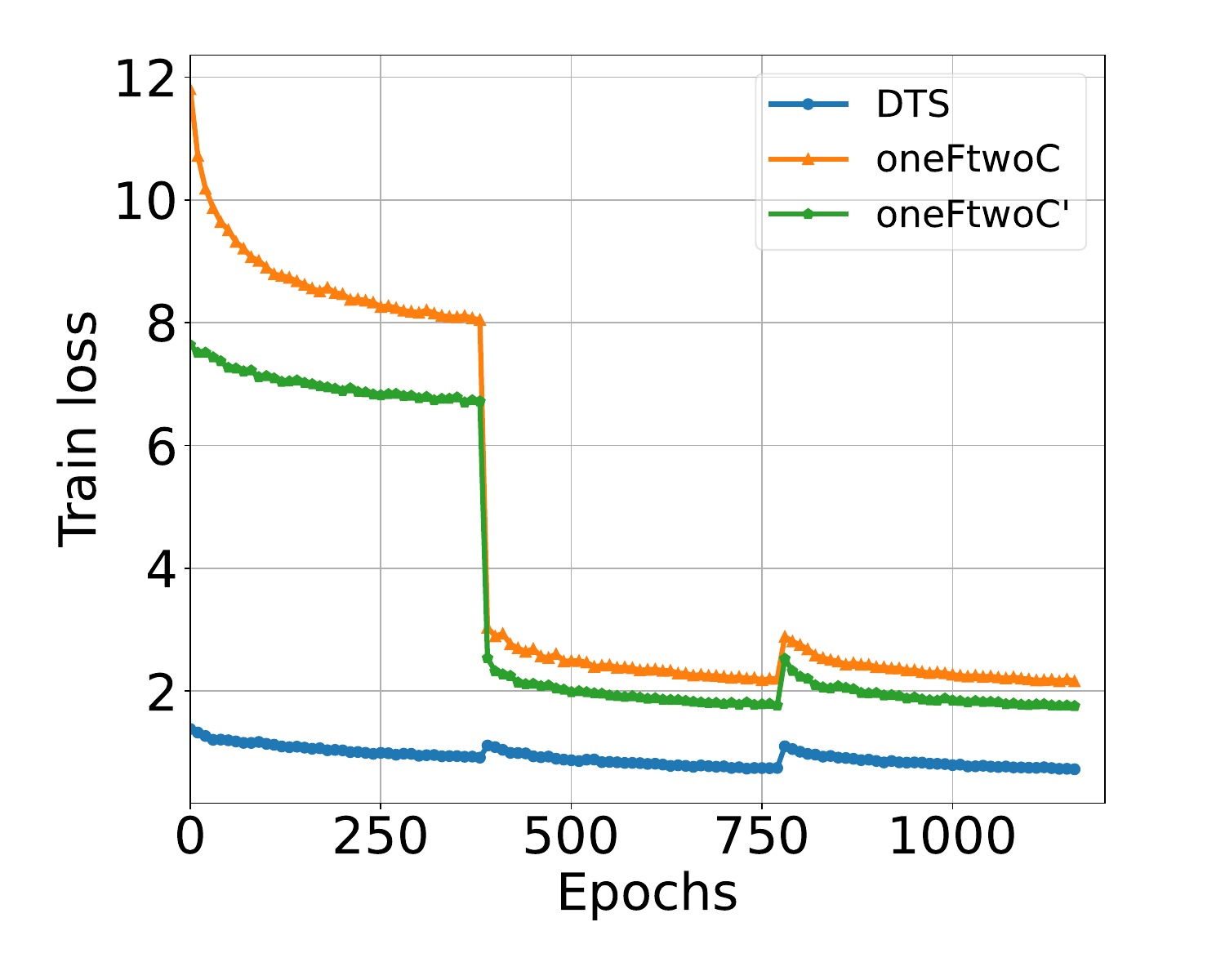}
    }
    \subfloat[train loss on SVHN]
    {
        \label{fig: train_loss_svhn_0.3}\includegraphics[width=0.32\textwidth]{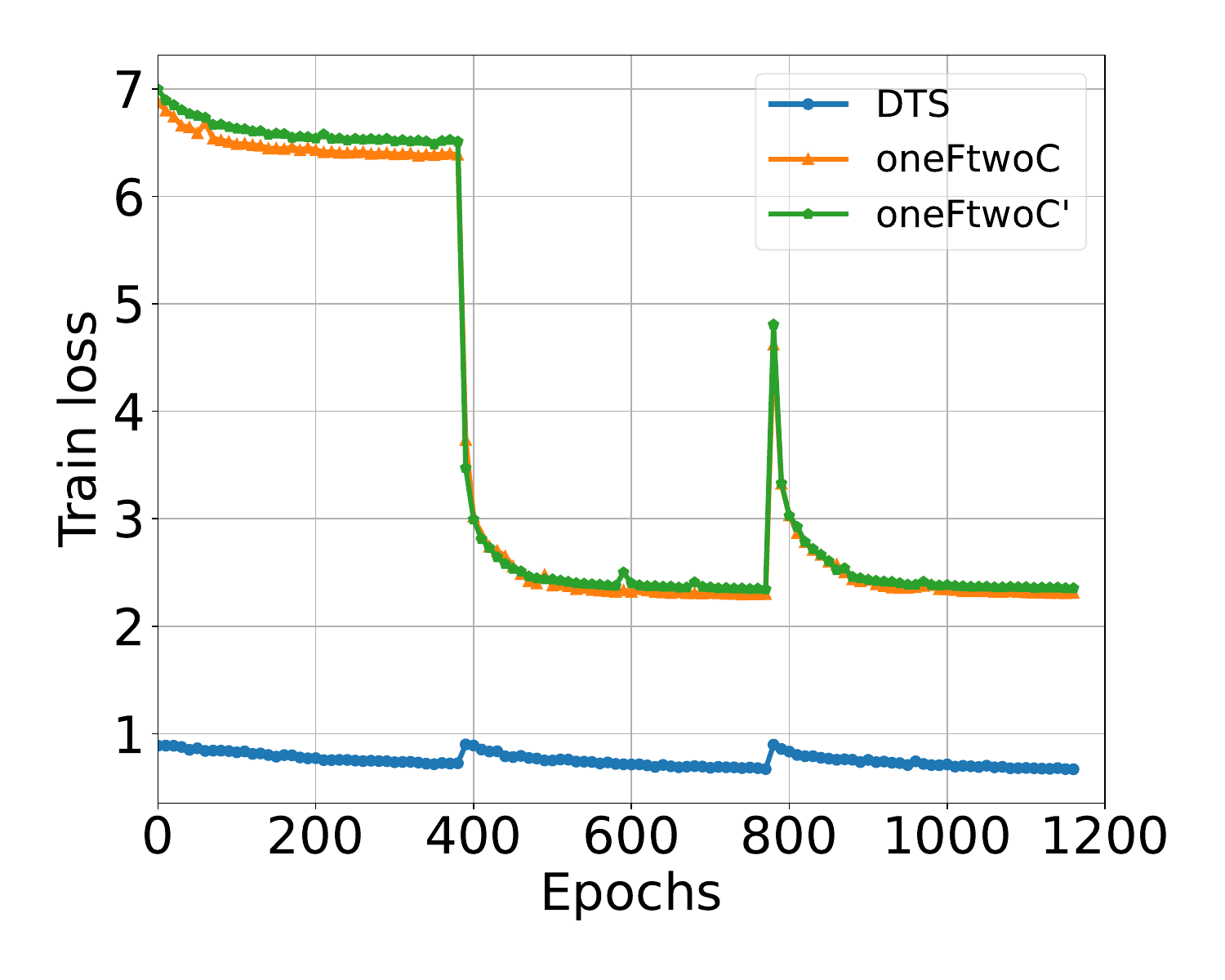}
    }
        \subfloat[train loss on CIFAR-100]
    {
        \label{fig: train_loss_cifar100_0.3}\includegraphics[width=0.32\textwidth]{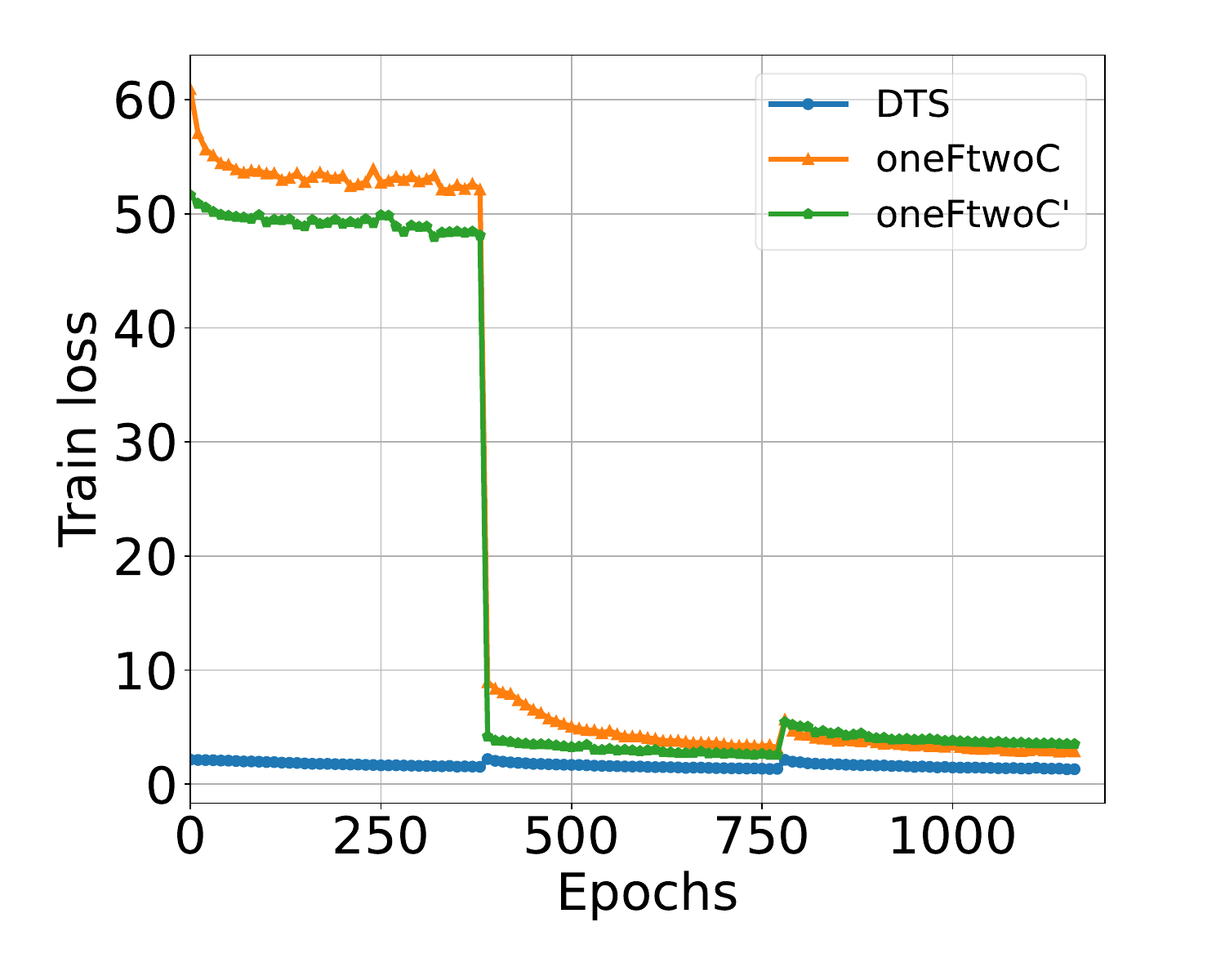}
    }
\caption{Comparison of train loss among DTS, oneFtwoC, and oneFtwoC{\textquotesingle} on CIFAR-10, SVHN and CIFAR-100 datasets with a class mismatch ratio of 0.3.}
\label{fig of loss and acc}
\end{figure*}

\begin{figure*}[t]
\centering
    \subfloat[test acc on CIFAR-10]
    {
        \label{fig: test_acc_cifar10_0.3}\includegraphics[width=0.32\textwidth]{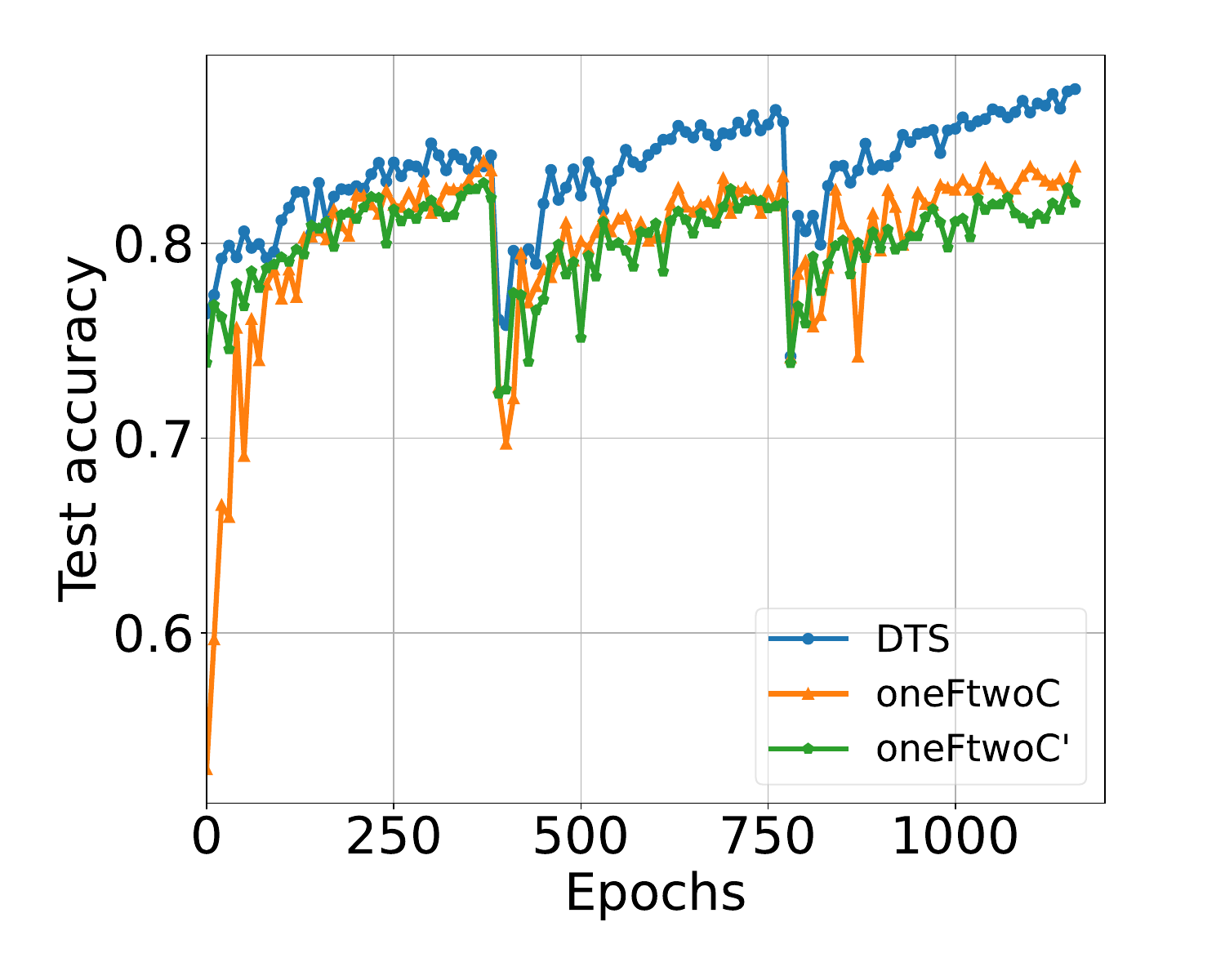}
    }
    \subfloat[test acc on SVHN]
    {
        \label{fig: test_acc_svhn_0.3}\includegraphics[width=0.32\textwidth]{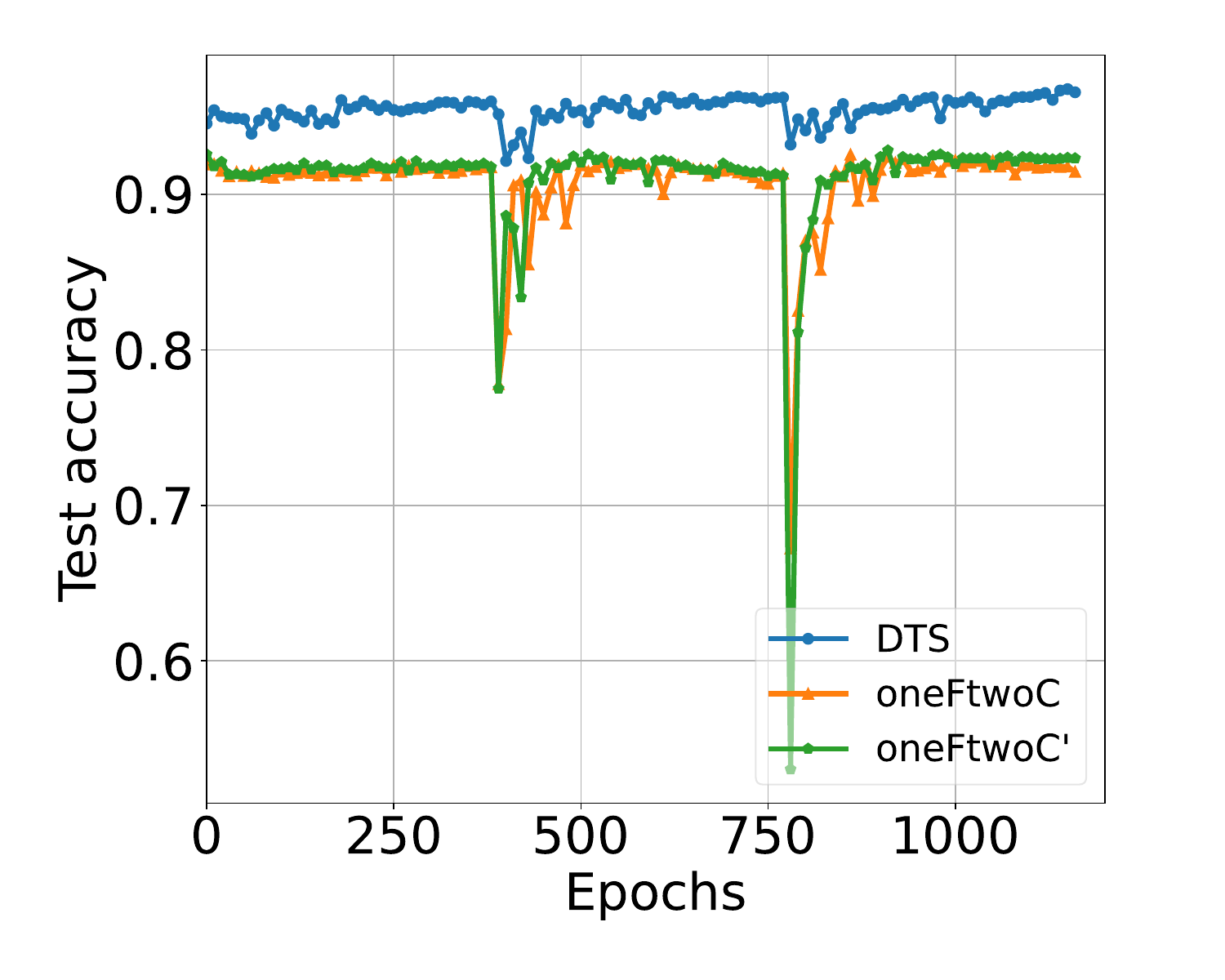}
    }
    \subfloat[test acc on CIFAR-100]
    {
        \label{fig: test_acc_cifar100_0.3}\includegraphics[width=0.32\textwidth]{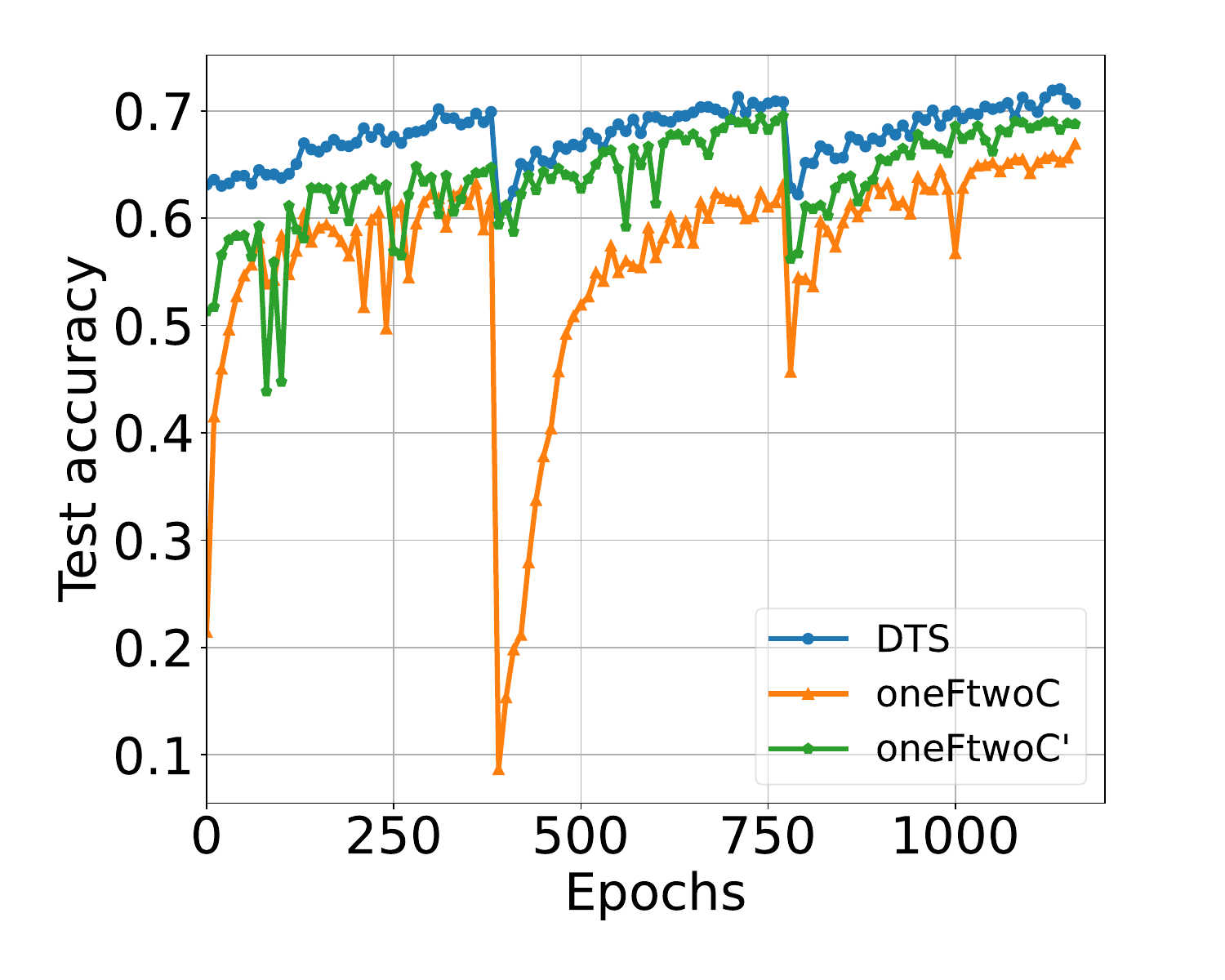}
    }
\caption{Comparison of test accuracy among DTS, oneFtwoC, and oneFtwoC{\textquotesingle} on CIFAR-10, SVHN and CIFAR-100 datasets with a class mismatch ratio of 0.3.}
\label{fig of loss and acc2}
\end{figure*}

\subsection{Compared to One Model with Two Classifiers}
To better validate the effectiveness and superiority of our proposed DTS, we also conducted experiments on two simplified versions of DTS, namely \textbf{oneFtwoC} and \textbf{oneFtwoC\textquotesingle}. Specifically, oneFtwoC refers to merging the two teacher-student frameworks of DTS into one, aiming to train a model that simultaneously possesses a $K$-classifier and a ($K$+1)-classifier, with soft-weighting operations between the two classifiers; while oneFtwoC{\textquotesingle} builds upon oneFtwoC by adding a projection head to the ($K$+1)-classifier, as it theoretically can enhance the model's performance. As illustrated in Figure \ref{fig of loss and acc} and Figure \ref{fig of loss and acc2}, we present the trends in train loss and test accuracy of the three models on the CIFAR-10, SVHN and CIFAR-100 datasets, with the class mismatch ratio set to 0.3.

In Figure \ref{fig of loss and acc}, we observed that the losses for oneFtwoC and oneFtwoC{\textquotesingle} plateau after a certain point, which is attributable to the inherent conflict between conducting $K$-classification and ($K$+1)-classification within a single model, leading to limited convergence during training. In Figure \ref{fig of loss and acc2}, it is evident that DTS consistently outperforms the oneFtwoC and oneFtwoC{\textquotesingle} methods. This advantageously supports our viewpoint that employing two classifiers directly within a single model can lead to conflicts, resulting in suboptimal training outcomes. Therefore, we proposed DTS, which utilizes two teacher-student frameworks to address the safe SSL issue. DTS significantly mitigates optimization conflicts between tasks by separating them, thereby facilitating the effective utilization of unlabeled data.

\section{Conclusion}

In this paper, we first study the prevailing training strategies of current safe SSL methods, which involve training a single model to perform two tasks: classification for seen classes and detection for unseen classes. We unveil the potential pitfalls of this approach, where the concurrent tasks may interfere with each other, leading to suboptimal model optimization. Building on this insight, we introduce the Diverse Teacher-Students (DTS) framework, which orchestrates the collaborative training of two models to independently handle these two tasks. Specifically, DTS employs a novel uncertainty score to softly separate unseen-class samples within unlabeled data, creating supervisory signals for the ($K$+1)-th class involved in training. Besides, the improvement of DTS in terms of AUROC, brought about by the soft-weighting module, is also noteworthy. Furthermore, DTS demonstrates exceptional performance without the need for additional prior knowledge of hyperparameters, making the task simpler, more feasible, and convenient to apply. We hope that our findings will offer fresh perspectives and insights into the optimal utilization of unlabeled data within the safe SSL domain. Additionally, given the rapid development of LLMs, we anticipate that the integration of safe SSL with LLMs will drive more intelligent and secure data-driven applications, significantly enhancing the accuracy and safety of models when handling large-scale unlabeled data.

\section*{Acknowledgment}
This work is supported by the National Natural Science Foundation of China (62176139, 61876098, 62202270), in part by the Major Basic Research Project of Natural Science Foundation of Shandong Province (ZR2021ZD15), in part by the Natural Science Foundation of Shandong Province, China (ZR2021QF034), in part by the Shandong Excellent Young Scientists Fund (Oversea) (2022HWYQ-044), and the Taishan Scholar Project of Shandong Province (tsqn202306066).

\bibliography{main}

\end{document}